\documentclass[11pt,a4paper]{article}
\usepackage{times,latexsym}
\usepackage{url}
\usepackage[T1]{fontenc}

\usepackage[dvipsnames,table]{xcolor}
\usepackage{graphicx}
\usepackage{amsmath}
\usepackage{booktabs}
\usepackage{tabularx}
\usepackage{wrapfig}
\usepackage{array}
\usepackage{appendix}
\usepackage{multirow}
\usepackage{hyperref}
\usepackage{longtable}
\usepackage{caption}
\usepackage{algorithm}
\usepackage{algpseudocode}
\usepackage{afterpage}
\usepackage{svg}

\usepackage{colortbl}

\usepackage[acceptedWithA]{tacl2021v1}

\usepackage{xspace,mfirstuc,tabulary}

\newif\iftaclinstructions
\taclinstructionsfalse 
\iftaclinstructions
\renewcommand{\confidential}{}
\renewcommand{\anonsubtext}{(No author info supplied here, for consistency with
TACL-submission anonymization requirements)}
\newcommand{\instr}
\fi

\iftaclpubformat 

\else

\fi

\title{OrderSum: Semantic Sentence Ordering for Extractive Summarization}

\author{
  Taewan Kwon$^1$
  \\
  \And
  Sangyong Lee$^2$
  \\
}
\date{}
\begin{document}
\maketitle


\begin{abstract}

There are two main approaches to recent extractive summarization: the sentence-level framework, which selects sentences to include in a summary individually, and the summary-level framework, which generates multiple candidate summaries and ranks them. Previous work in both frameworks has primarily focused on improving which sentences in a document should be included in the summary. However, the sentence order of extractive summaries, which is critical for the quality of a summary, remains underexplored. In this paper, we introduce OrderSum, a novel extractive summarization model that semantically orders sentences within an extractive summary. OrderSum proposes a new representation method to incorporate the sentence order into the embedding of the extractive summary, and an objective function to train the model to identify which extractive summary has a better sentence order in the semantic space. Extensive experimental results demonstrate that OrderSum obtains state-of-the-art performance in both sentence inclusion and sentence order for extractive summarization. In particular, OrderSum achieves a ROUGE-L score of 30.52 on CNN/DailyMail, outperforming the previous state-of-the-art model by a large margin of 2.54\footnote{Our code and the model parameters are available at \url{https://github.com/Espresso-AI/OrderSum}.}.

\end{abstract}


\section{Introduction}

In the field of text summarization, there are two main techniques: extractive summarization \citep{cheng-lapata-2016-neural, nallapati_summarunner_2017, zhong-etal-2020-extractive} and abstractive summarization \citep{rush-etal-2015-neural, nallapati-etal-2016-abstractive, liu-liu-2021-simcls}. Abstractive summarization creates a natural summary by generating text token by token, while extractive summarization provides a summary with more accurate information by extracting sentences directly from the document.

There are two main approaches to neural extractive summarization for a single document: 1) the sentence-level framework \citep{nallapati_summarunner_2017, zhang-etal-2018-neural, zhou-etal-2018-neural-document, liu-lapata-2019-text} and 2) the summary-level framework \citep{narayan-etal-2018-ranking, zhong-etal-2020-extractive, an-etal-2022-colo, zhang-etal-2023-diffusum}. The sentence-level framework defines extractive summarization as an individual sentence selection problem, determining whether each sentence in a document should be included in the summary. However, the sentence-level framework often produces summaries that contain only general sentences or repeat important but similar sentences \citep{narayan-etal-2018-ranking, zhong-etal-2020-extractive}. The summary-level framework overcomes this limitation by defining extractive summarization as a summary ranking problem rather than a sentence selection problem. The main idea of the summary-level framework is to generate a set of candidate summaries consisting of different sentences, and then rank them to select the best summary. By considering sentence composition at the entire summary level rather than sentence by sentence, this approach enables each sentence in the summary to convey different, specific information \citep{narayan-etal-2018-ranking, zhong-etal-2020-extractive}.

Previous work in both frameworks has primarily focused on improving which sentences to include in the summary, or in other words, sentence inclusion. However, to the best of our knowledge, the importance of sentence order in summaries has not been highlighted since the era of graph-based extractive summarization \citep{mihalcea-tarau-2004-textrank, erkan_lexrank_2004}. The sentence order of a text plays a crucial role not only in readability but also in its meaning \citep{ijcai2019p0748, 10.5555/3504035.3504683}. Table 1 illustrates how the meaning of summaries can change simply by altering the sentence order. Although both summaries contain the same sentences, the first summary conveys similar content to the reference summary, while the second does not because of the difference in sentence order. This shows the necessity of considering sentence order as a semantic factor in extractive summarization.

\begin{table}[t!]
  \centering
  \setlength{\tabcolsep}{6pt}
  \renewcommand{\arraystretch}{1.4}
  \small
  \begin{tabular}{p{0.95\columnwidth}}
    \toprule
    \textbf{Reference Summary} \\
    \hline
    \small An international team of scientists introduced a breakthrough treatment in their research paper that could save millions of lives suffering from the side effects of cancer treatment. The announcement brought hope and increased investment to the research team, even though human trials have not yet been conducted. \\

    \toprule
    \textbf{Example 1} \\
    \hline
    \small \textcolor{OliveGreen}{\textbf{[0]} The international team of scientists published a paper on a cancer treatment that targets only cancer cells with very high precision, potentially minimizing the side effects of existing treatments.} \textcolor{RoyalBlue}{\textbf{[1]} At the medical conference, they said the treatment is still in the animal testing phase, and its human efficacy is unknown.} \textcolor{RedOrange}{\textbf{[2]} Despite this, the announcement sparked massive investment in the team.} \\
    
    \toprule
    \textbf{Example 2} \\
    \hline
    \small \textcolor{OliveGreen}{\textbf{[0]} The international team of scientists published a paper on a cancer treatment that targets only cancer cells with very high precision, potentially minimizing the side effects of existing treatments.} \textcolor{RedOrange}{\textbf{[2]} Despite this, the announcement sparked massive investment in the team.} \textcolor{RoyalBlue}{\textbf{[1]} At the medical conference, they said the treatment is still in the animal testing phase, and its human efficacy is unknown.} \\
    
    \bottomrule 
  \end{tabular} 
  \normalsize
  \caption{Examples of extractive summaries that share the same sentences but have different orders. The first summary is a human-written reference summary that does not directly contain a sentence in the original document. To be semantically aligned with the reference summary, the sentences must be ordered as in Example 1.}
  \label{table:1}
\end{table}

In this paper, we present OrderSum, a novel extractive summarization model that semantically orders sentences within an extractive summary. The main ideas of OrderSum are as follows:
\begin{enumerate}
    \item OrderSum introduces a new representation method to semantically incorporate the sentence order into the embedding of the extractive summary.

    \item OrderSum proposes an objective function to train the model to identify which extractive summary has a better sentence order in the semantic space, based on the summary-level framework.
\end{enumerate}
Consequently, OrderSum generates a set of extractive candidate summaries with different sentence orders and identifies the summary closest to the document in the semantic space by considering both sentence inclusion and sentence order.

We evaluate our model on four datasets. Results show that OrderSum achieves state-of-the-art performance on ROUGE-L \citep{lin-2004-rouge}, the metric for evaluating sentence order of the entire summary. On CNN/DailyMail \citep{NIPS2015_afdec700}, OrderSum achieves a ROUGE-L score of 30.52, outperforming the previous state-of-the-art model by a margin of 2.54. We found that while existing summary-level framework models improve sentence inclusion from the sentence-level framework, they do not achieve meaningful improvements in sentence order. To investigate this, we train the summary-level framework model and compare its validation graphs during training against those of our model. Finally, since large language models (LLMs) outperform task-specific models in various natural language processing fields \citep{radford2019language, NEURIPS2020_1457c0d6}, we evaluate the recent LLM performance on the extractive summarization task.


\section{Related Work}

\subsection{Sentence-level Framework}

Single document extractive summarization is the task of obtaining a summary $S$ corresponding to a subset of a source document $D = \{s_1, \ldots, s_n\}$, where $s$ denotes a sentence and $n$ is the number of sentences in the document. The sentence-level framework is a standard approach to neural extractive summarization that determines whether each sentence should be included in the summary or not. The sentence-level framework trains the model using an objective function based on maximum likelihood estimation (MLE) for individual sentences \citep{nallapati_summarunner_2017, liu-lapata-2019-text}:
\begin{equation}
\label{eq1}
\theta^* = \underset{\theta}{\text{argmax}} \underset{i}\sum \, log \, p_{f_{\theta}}(s_i \in D),
\end{equation}
where $f$ is the model with parameters $\theta$, $\theta^*$ is the optimized parameters of $f$, and $s_i$ denotes the $i$-th sentence in the document $D$. $p_{f_{\theta}}$ refers to the probability of each sentence being included in the summary. During inference, the summary $S$ is determined by sorting the sentences in descending order of $p_{f_{\theta}}$ and selecting the top-$k$ sentences.

However, a summary consisting only of sentences with the highest $p_{f_{\theta}}$ does not always guarantee a representative summary for the document, because the summary may contain only general sentences or repeat important but similar sentences \citep{narayan-etal-2018-ranking, zhong-etal-2020-extractive}. In contrast, a good summary covers the entire content of the document. Accordingly, summarization metrics \citep{lin-2004-rouge, Zhang*2020BERTScore:, NEURIPS2021_e4d2b6e6, zhao-etal-2019-moverscore} compare the reference summary $S^*$ and the model summary $S$ as a whole, not sentence by sentence. The limitation of a sentence-level framework is that the model parameters are optimized for the sentence-level objective function, in contrast to the evaluation at the summary level, which is called the inherent gap problem \citep{narayan-etal-2018-ranking, zhong-etal-2020-extractive}.


\subsection{Summary-level Framework}

The summary-level framework was proposed to address the inherent gap problem of the sentence-level framework. The main idea of the summary-level framework is to generate a set of candidate summaries, $\mathcal{C} = \{C_1, \ldots, C_m\}$, consisting of different sentences, and to employ a model, called a reranker, that obtains the ranking of the candidates to find the best one $\hat{C}$ at the summary level. By training the reranker with the summary-level objective function, the summary-level framework overcomes the inherent gap between training and evaluation.

The summary-level framework generally proceeds in four steps \citep{narayan-etal-2018-ranking, zhong-etal-2020-extractive, an-etal-2022-colo}:

\noindent
\textbf{Extracting Key Sentences}: The first step of the summary-level framework is to extract the key sentences that serve as the elements of the candidate summaries. For this step, a sentence-level extractive summarization model, called an extractor, is employed. The extractor selects the top-$k$ sentences with the highest $p_{f_{\theta}}$ as the key sentences.

\noindent
\textbf{Generating Candidate Summaries}: The second step is to create a set of candidate summaries $\mathcal{C}$ containing different sentences. The $\mathcal{C}$ is mainly generated by a combination of the key sentences $\{{}_{k} C_r \,|\, r \in \mathcal{N},\ r \leq k\}$, where $r$ is the number of sentences in each candidate summary, and $\mathcal{N}$ is the set of possible values for $r$.

\noindent
\textbf{Embedding Candidate Summaries}: Each candidate summary is represented as a single embedding vector before reranking. The candidate summaries are embedded by the reranker, which uses RNN-based layers \citep{narayan-etal-2018-ranking}, transformer layers, or another encoder model \citep{zhong-etal-2020-extractive} such as BERT \citep{devlin-etal-2019-bert}. Recently, DiffuSum \citep{zhang-etal-2023-diffusum} introduced a diffusion model for embedding candidate summaries.

\noindent
\textbf{Reranking Candidate Summaries}: To address the inherent gap between training and evaluation, the reranker is trained using a summary-level objective function. Early studies applied reinforcement learning with rewards based on evaluation metrics \citep{narayan-etal-2018-ranking, bae-etal-2019-summary, dong-etal-2018-banditsum}. Semantic matching is a prominent approach that uses contrastive learning \citep{oord2019representationlearningcontrastivepredictive} to find the candidate summary most similar to a document in the semantic space \citep{zhong-etal-2020-extractive, liu-liu-2021-simcls, an-etal-2022-colo, zhang-etal-2023-diffusum}. The main idea behind these approaches is to train the reranker to identify the candidate summary closest to the reference summary $S^*$ during training by including summary-level evaluation metrics in the objective function. Depending on which evaluation metric is used, the priority of the candidate summaries is determined during training.

To the best of our knowledge, previous studies in both the sentence-level and summary-level frameworks have not focused on sentence order at the entire summary level. Sentence-level framework models arrange sentences in the summary in descending order of $p_{f_{\theta}}$, determined at the sentence level. On the other hand, summary-level framework models do not generate candidate summaries that share the same sentences but have different orders, nor do they propose an objective function to train the sentence order. This makes the summary-level framework improve sentence inclusion but still suffer from the inherent gap problem in sentence order.


\section{Model Architecture}

\begin{figure*}[t!]
  \centering
  \includegraphics[width=\linewidth]{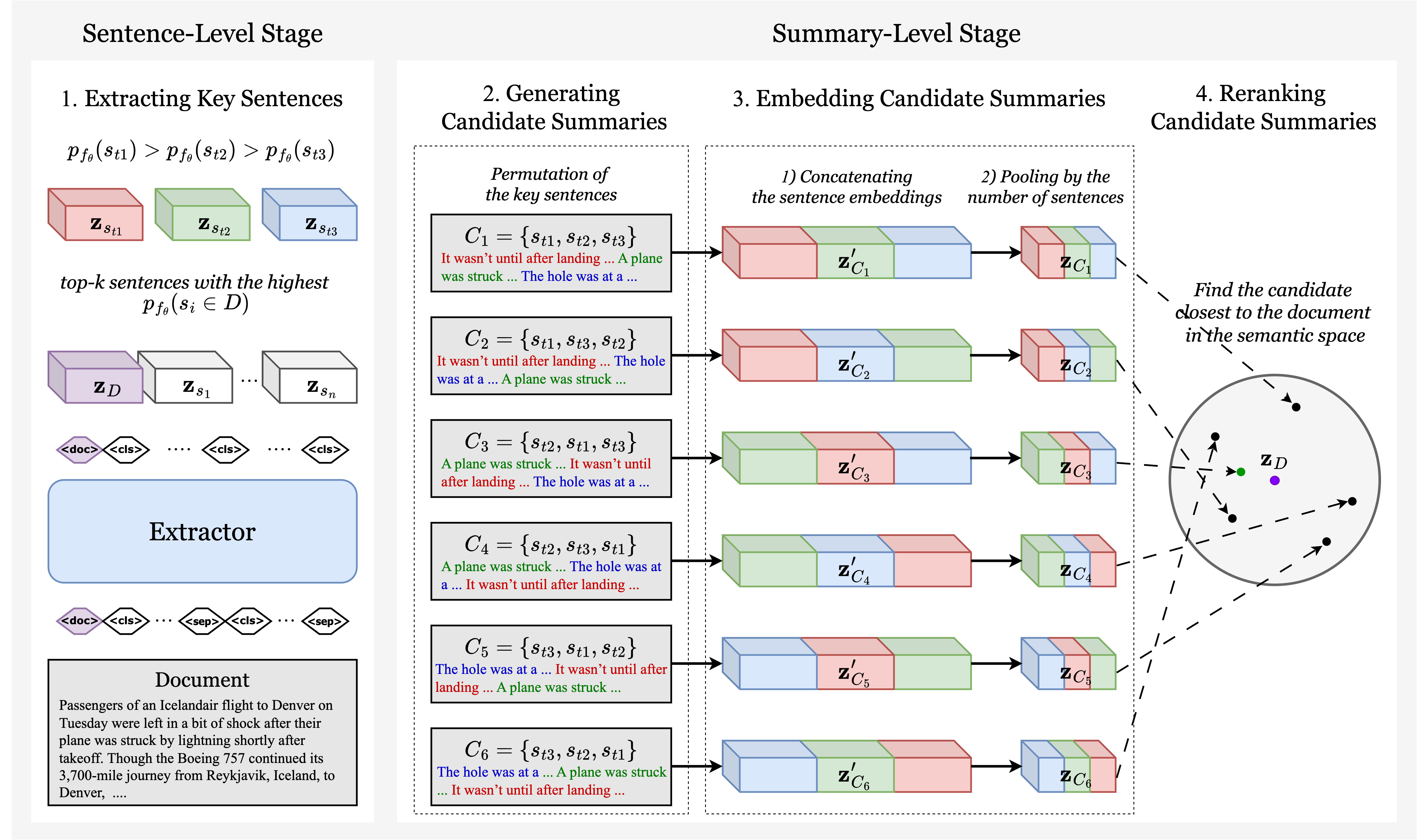}
  \caption{The structure of OrderSum is based on the summary-level framework. The first step is to extract key sentences using the extractor. In the second step, OrderSum generates the set of candidate summaries with different sentence orders. In the next step, OrderSum obtains the candidate summary embeddings representing the sentence order of each summary. Finally, OrderSum trains the sentence order using a new objective function in the semantic space.}
  \label{fig:fig1}
\end{figure*}

In this section, we describe the structure of OrderSum in four steps of the summary-level framework. The goal of OrderSum is to obtain an extractive summary in which the sentences are semantically ordered. In order to achieve this, OrderSum transforms the steps of generating and embedding candidate summaries, and proposes a new objective function to train the sentence order in the semantic space.


\subsection{Extracting Key Sentences}

The first step of OrderSum is to extract the key sentences that serve as the elements of the candidate summary set $\mathcal{C}$ using the extractor. Before being fed to the extractor, the document is separated into individual sentences during tokenization using special tokens ‘<cls>’ and ‘<sep>’, which denote the start and end of each sentence, respectively \citep{liu-lapata-2019-text}. Another special token ‘<doc>’ is added to the beginning of the document \citep{an-etal-2022-colo}. The token sequence of the document is then passed to the extractor, which is a sentence-level framework model. Through the extractor, the embedding of each ‘<cls>’ token $\mathbf{z}_{s_i}$ represents each sentence, and the embedding of the ‘<doc>’ token $\mathbf{z}_D$ represents the entire document \citep{liu-lapata-2019-text, an-etal-2022-colo}. Finally, each $\mathbf{z}_{s_i}$ is processed using a sigmoid function to compute the probability $p_{f_{\theta}}$ of each sentence being included in the summary. The top-$k$ sentences with the highest $p_{f_{\theta}}$ are selected as the key sentences to create the candidate summaries.


\subsection{Generating Candidate Summaries}

The second step is to generate the set of candidate summaries $\mathcal{C}$ using the key sentences. The main difference from the existing methods is to create the candidates that vary not only in sentence inclusion but also in sentence order. To achieve this, $\mathcal{C}$ is defined as a permutation:
\begin{equation}
\label{eq2}
\mathcal{C} = \{{}_{k} P_r \,|\, r \in \mathcal{N},\ r \leq k\},
\end{equation}
where $r$ is the number of sentences in each candidate summary, and $\mathcal{N}$ is the set of possible values for $r$. As shown in Figure 1, this generates subsets of candidate summaries that share the same sentences but differ in sentence order.


\subsection{Embedding Candidate Summaries}

The following step is to represent the candidate summaries as embedding vectors in the semantic space. To incorporate the sentence order into the candidate summary embeddings, OrderSum represents the candidate summaries in two steps:
\begin{enumerate}
    \item Concatenating sentence embeddings along the sequence dimension.

    \item Pooling into the same channel size to represent in the same semantic space.
\end{enumerate}
As shown in Figure 1, OrderSum first concatenates sentence embeddings in the order in which they appear in a candidate summary along the sequence dimension:
\begin{equation}
\label{eq3}
\mathbf{z}_{C_j}^\prime = \left[\,\mathbf{z}_{s_t} \,|\, s_t \in C_j \,\right],
\end{equation}
where $s_t$ refers to a sentence in a candidate summary $C_j$, $\mathbf{z}_{s_t}$ is the embedding vector of $s_t$, and $\mathbf{z}_{C_j}^\prime$ is the intermediate embedding vector of $C_j$. In this step, the sentence order of each candidate summary is directly represented in $\mathbf{z}_{C_j}^\prime$, and candidate summaries containing the same sentences but with different sentence orders obtain different embeddings based on the sentence order.

The intermediate embedding vectors of the candidate summaries are then pooled to the same channel size as the document embedding $\mathbf{z}_D$. In order to compute the similarity between the document and the candidate summaries in the semantic space, the embeddings of the document and the candidate summaries must have the same channel size. However, during concatenation, the channel size of the candidate summary embeddings depends on the number of sentences, which varies for each candidate summary. To address this problem, OrderSum proposes sentence-number pooling, where the window size and stride size are set to match the number of sentences in each candidate summary:
\begin{equation}
\label{eq4}
\mathbf{z}_{C_j} = \left[\,\frac{1}{r} \sum_{\gamma=0}^{r-1} \mathbf{z}_{C_j}^\prime \left[r\delta + \gamma\right] \,\right]_{\delta=0}^{d - 1}.
\end{equation}
In Eq. 4, $r$ is the number of sentences in $C_j$, and $d$ is the channel size of $\mathbf{z}_D$. $\mathbf{z}_{C_j}^\prime \left[r\delta + \gamma\right]$ is the $(r\delta + \gamma)$-th scalar value of the intermediate embedding vector $\mathbf{z}_{C_j}^\prime$. The outer brackets $\left[\cdot\right]_{\delta=0}^{d - 1}$ denote average pooling along the sequence dimension. $\mathbf{z}_{C_j}$ is the final embedding vector of $C_j$. As a result, OrderSum maps all candidate summaries $\mathcal{C}$ with different sentence inclusion and sentence order in the same semantic space as the document.

As shown in Figure 1, candidate summaries that share the same sentences but have different sentence orders are positioned differently in the semantic space. It enables the reranker to learn the sentence order, unlike the case where candidate summaries with different sentence orders are indistinguishable in the semantic space.


\subsection{Reranking Candidate Summaries}

OrderSum follows a semantic matching approach based on the cosine similarity in the semantic space. To obtain the final candidate summary $\hat{C}$, OrderSum computes the cosine similarity of the candidate summary embeddings to the document embedding $\mathbf{z}_D$ and selects the candidate with the highest similarity value.
\begin{equation}
\label{eq5}
\hat{C} = \underset{C_j \in \,\mathcal{C}}{\text{argmax}}\;\text{cos}(\mathbf{z}_D, \mathbf{z}_{C_j}).
\end{equation}


\subsection{Objective Function for Training OrderSum}

The main concept of training OrderSum is to include a metric that evaluates sentence order at the entire summary level in the objective function. OrderSum integrates this concept with contrastive learning to semantically train sentence order.

The final objective function of OrderSum is a multi-task learning function:
\begin{equation}
\label{eq6}
\mathcal{L} = \mathcal{L}_{sent} + \mathcal{L}_{sum},
\end{equation}
consisting of objective functions for sentence extraction $\mathcal{L}_{sent}$ and summary reranking $\mathcal{L}_{sum}$.

First, $\mathcal{L}_{sent}$ is the MLE-based objective function of the sentence-level framework:
\begin{equation}
\label{eq7}
\begin{split}
\mathcal{L}_{sent} &= -\frac{1}{n} \ \underset{i}{\sum}\,(y_i \, log \,  p_{f_{\theta}}(s_i) \\
&+ (1-y_i) \, log(1-p_{f_{\theta}}(s_i))),
\end{split}
\end{equation}
where $n$ is the number of sentences in the document, $s_i$ and $y_i$ are the $i$-th sentence in the document and its label, and $p_{f_{\theta}}$ is the probability of each sentence being included in the summary, obtained by the extractor. Since OrderSum uses sentence embeddings directly from the extractor to obtain candidate summary embeddings, the extractor and the reranker share the same encoder parameters, like CoLo \citep{an-etal-2022-colo}. To prevent degradation in sentence extraction during training on summary reranking, $\mathcal{L}_{sent}$ is included in $\mathcal{L}$.

The objective function $\mathcal{L}_{sum}$ for summary reranking is a triplet ranking loss \citep{Schroff_2015_CVPR} for contrastive learning:
\begin{equation}
\label{eq8}
\begin{split}
\mathcal{L}_{sum} &= \underset{j}{\sum}\underset{k > j}{\sum}\;\text{max}(0,\; \text{cos}(\mathbf{z}_D, \mathbf{z}_{C_k}) \\
&- \text{cos}(\mathbf{z}_D, \mathbf{z}_{C_j}) + \lambda_{jk}),
\end{split}
\end{equation}
where $j$ and $k$ refer to a pair of indices in the set of the candidate summaries. $\mathbf{z}_{D}$ is the document embedding, and $\mathbf{z}_{C_j}$, $\mathbf{z}_{C_k}$ are the embeddings of the indexed candidate summaries. $\text{cos}$ denotes the cosine similarity. $\lambda_{jk}$ is the margin between the two distances from $\mathbf{z}_{D}$ to $\mathbf{z}_{C_j}$ and $\mathbf{z}_{C_k}$ on the semantic space, computed as $\lambda \, (k - j)$, where $\lambda$ is 0.01. In Eq. 8, $D$, $C_j$, and $C_k$ correspond to the anchor, positive sample, and negative sample of the triplet ranking loss, respectively. The model learns that the candidate summary $C_j$ is relatively closer to the document $D$ than $C_k$ in the semantic space \citep{zhong-etal-2020-extractive}.

During training, it must be determined which candidate summaries should be provided to $\mathcal{L}_{sum}$ as positive and negative samples. The ranking function $H$ is defined to rank the candidate summaries and generate pairs of positive and negative samples from the ranking of the candidates \citep{zhong-etal-2020-extractive}. First, the ranking function computes the ranking score between the reference summary $S^*$ and each candidate summary during training. The candidate summaries are then sorted based on the ranking scores. Finally, every pair of $C_j$ and $C_k$ that satisfies $H(C_j) > H(C_k)$ is obtained from the sorted ranking of the candidates. Consequently, the ranking function $H$ decides the priority of the candidate summaries for the model to learn.

OrderSum proposes a new ranking function that ranks the candidate summaries according to both sentence inclusion and sentence order:
\begin{equation}
\label{eq9}
\begin{split}
H(C_j) &= \text{ROUGE-1}(S^*, C_j) \\
&+ \text{ROUGE-2}(S^*, C_j) \\
&+ \text{ROUGE-L}(S^*, C_j).
\end{split}
\end{equation}
In Eq. 9, ROUGE-L is added as a metric for training sentence order. To achieve this, the ROUGE-L term must evaluate the sentence order of a candidate summary $C_j$ against the reference summary $S^*$ at the entire summary level. We find that this condition is satisfied when the longest common subsequence (LCS) between the two summaries is calculated by treating each summary as a single complete text while computing ROUGE-L \citep{lin-2004-rouge} (see details in Appendix 1). By integrating the new ranking function with contrastive learning, the model learns that a candidate summary with a sentence order closer to the reference summary is also semantically closer to the document. Consequently, OrderSum is trained to rank the candidate summaries by considering both sentence inclusion and sentence order at the entire summary level.


\subsection{Anchor Candidate Sampling}

A critical problem in training OrderSum is the large number of candidate summaries due to using permutation to generate the set of candidate summaries. Because Eq. 8 is computed for every pair of candidate summaries \citep{zhong-etal-2020-extractive, an-etal-2022-colo}, the training cost grows as $O(n^2)$ as the number of candidates increases. To address this problem, we propose anchor candidate sampling, an effective method to sample the candidate summaries used for training on summary reranking (see Appendix 2). The experiments in Appendix 2 show that anchor candidate sampling dramatically accelerates training with minimal performance degradation.


\section{Experimental Setup}

\begin{table*}[t!]
  \centering
  \setlength{\tabcolsep}{10.27pt} 
  \renewcommand{\arraystretch}{1.35} 
  \begin{small}
  \begin{tabular}{lccccccccc}
    \toprule
      \multirow{2}{*}{\textbf{Datasets}} & \multicolumn{3}{c}{\textbf{\# data}} & \multicolumn{2}{c}{\textbf{\# tokens}} & \multicolumn{4}{c}{\textbf{\# candidates}} \\
      \cmidrule(lr){2-4} \cmidrule(lr){5-6} \cmidrule(lr){7-10}
      & \textbf{train} & \textbf{valid} & \textbf{test} & \textbf{$D$} & \textbf{$S^*$} & \textbf{$k$} & \textbf{$\mathcal{N}$} & \textbf{${}_{k} C_r$} & \textbf{${}_{k} P_r$} \\
      \midrule
      \textbf{CNN/DM} &  287,084 &  13,367 & 11,489 &  766.1 &  58.2 & 5 & 2, 3 & 20 & 80 \\
      \textbf{XSum} & 203,028 &  11,273 &  11,332 &  430.2 &  23.3 & 5 & 1, 2 & 15 & 25 \\
      \textbf{WikiHow} & 168,126 & 6,000 & 6,000 & 580.8 & 62.6 & 5 & 3, 4, 5 & 16 & 300 \\
      \textbf{PubMed} & 83,233 &  4,946 & 5,025 & 444.0 & 209.5 & 8 & 6, 7 & 36 & 60,480 \\
      \bottomrule
  \end{tabular}
  \end{small}
  \caption{Statistics of each dataset. $\textbf{train}$, $\textbf{valid}$, and $\textbf{test}$ under $\textbf{\# data}$ refer to the size of the training, validation, and test sets of each dataset. $D$ and $S^*$ under $\textbf{\# tokens}$ refer to the average number of tokens in the source document and the reference summary. Below $\textbf{\# candidates}$, $k$ is the number of key sentences from the extractor, and $\mathcal{N}$ is the set of possible values for the number of sentences in each candidate summary, $r$. ${}_{k} C_r$ and ${}_{k} P_r$ are the numbers of candidate summaries generated by combination and permutation, respectively.}
  \label{table:2}
\end{table*}

\subsection{Datasets}
We conducted experiments on four datasets. Table 2 shows detailed statistics for each dataset.

\textbf{CNN/DailyMail} \citep{NIPS2015_afdec700} is a standard dataset for summarization, obtained from news articles. We used a non-anonymized version.

\textbf{XSum} \citep{narayan-etal-2018-dont} is a highly abstractive dataset obtained from BBC news articles. The summaries answer the question, "What is the article about?" in a single sentence.

\textbf{WikiHow} \citep{koupaee2018wikihow} is a knowledge-based dataset obtained from the website WikiHow. The summaries consist of subheadings from each paragraph, evenly distributed across the source document.

\textbf{PubMed} \citep{cohan-etal-2018-discourse} is a summarization dataset for the biomedical domain. We follow the settings of MatchSum \citep{zhong-etal-2020-extractive}, where the source document and reference summary correspond to the introduction and abstract sections of a paper, respectively.


\subsection{Evaluation}

We use the ROUGE scores \citep{lin-2004-rouge} as evaluation metrics during the experiments \footnote{We use f-measure of ROUGE-1, ROUGE-2, and ROUGE-L, following the baseline models in our experiments.}. In particular, we focus on ROUGE-L, which evaluates sentence order performance. There are two methods for evaluating summarization using ROUGE-L, depending on how the LCS between the reference summary and the model summary is measured. The first method is based on the normalized pairwise LCS \citep{radev2002evaluation}, which splits each summary into sentences and matches the sentence pairs with the maximum LCS \citep{lin-2004-rouge}. The second method calculates the LCS without splitting the summary into sentences, treating each summary as a single complete text \citep{lin-2004-rouge}. Appendix 1 provides a detailed explanation and experimental comparison of the two methods. It verifies that the first method does not compare the sentence order between two summaries because of sentence pair matching, whereas the second method evaluates the sentence order at the entire summary level. Consequently, we adopt the second method for calculating ROUGE-L in our experiments (it is denoted as $\text{ROUGE-L}_{full}$ in Appendix 1).

To set the upper bound for ROUGE-L evaluation, we propose Ordered ORACLE, an oracle summary that improves the sentence order of the existing oracle summary. An oracle summary is an extractive summary created to be closest to the abstractive reference summary from the given document \citep{lin-hovy-2003-potential, hirao-etal-2017-enumeration}. The oracle summary serves as an upper bound for evaluating extractive summarization. Previous work has mainly relied on the oracle summary proposed by SummaRuNNer \citep{nallapati_summarunner_2017}, which employs a greedy algorithm to iteratively select sentences from a document to maximize the ROUGE-1 and ROUGE-2 scores against the reference summary. However, this method does not provide an upper bound for ROUGE-L because ROUGE-L is excluded. To obtain Ordered ORACLE, which contains the same sentences as the existing oracle summary but achieves the highest ROUGE-L, we permute the sentences within the existing oracle summary and select the one with the highest ROUGE-L against the reference summary.


\subsection{Implementation Details}

The parameters of OrderSum are initialized with an extractor trained in the sentence-level framework \citep{an-etal-2022-colo}. We employ BERTSUM \citep{liu-lapata-2019-text} and BARTSUM \citep{lewis-etal-2020-bart} as initialization models. The input sequence length is set to 512 for both models, but we additionally use BARTSUM with an input sequence length of 1024 on CNN/DailyMail \citep{an-etal-2022-colo}.

Motivated by CoLo \citep{an-etal-2022-colo}, OrderSum uses sentence embeddings directly to obtain candidate summary embeddings, so that the extractor and the reranker share the same encoder parameters. To train the encoder for both sentence extraction and summary reranking, the training is divided into two phases. In the first phase, the model is trained using Eq. 7 to learn sentence extraction only. In the second phase, the model is trained using Eq. 6 to learn sentence extraction and summary reranking simultaneously.

The AdamW optimizer \citep{loshchilov2018decoupled} with $\beta_1 = 0.9$, $\beta_2 = 0.999$ is used for both training phases. The learning rate scheduler is:
\begin{equation}
\label{eq10}
lr = lr_0 \cdot \min\left(step^{-0.5},\; step \cdot warmup^{-1.5}\right),
\end{equation}
where $step$ denotes each step of training and $warmup$ is warmup steps of 10,000 \citep{NIPS2017_3f5ee243}. The constant $lr_0$ is set to $2e^{-3}$ and $1e^{-3}$ for the first and the second phases of the training, respectively. We train our model on a single Nvidia Tesla T4 and use gradient accumulation to ensure that models of different sizes are updated with the same batch size of 32.


\section{Results}

\begin{table}[t!]
  \centering
  \setlength{\tabcolsep}{6pt} 
  \renewcommand{\arraystretch}{1.5} 
  \small
    \begin{tabular}{lcccccc}
      \toprule
      \multicolumn{1}{c}{\textbf{}} & \textbf{R-1} & \textbf{R-2} & \cellcolor[gray]{0.9}\textbf{R-L}\\
      
      \hline
      \textbf{LEAD} & 40.14 & 17.55 & \cellcolor[gray]{0.9}25.08 \\
      \textbf{ORACLE} & 56.29 & 33.77 & \cellcolor[gray]{0.9}39.98 \\
      \textbf{Ordered ORACLE} & 56.29 & 33.82 & \cellcolor[gray]{0.9}44.11 \\
      
      \hline
      \textbf{BERTSUM-EXT} & 43.03 & 20.16 & \cellcolor[gray]{0.9}27.69 \\
      \textbf{BERTSUM} & 42.90 & 20.06 & \cellcolor[gray]{0.9}27.62 \\
      \textbf{BARTSUM} & 43.81 & 20.85 & \cellcolor[gray]{0.9}28.09 \\
      \textbf{BARTSUM 1024} & 44.09 & 21.09 & \cellcolor[gray]{0.9}28.34 \\
    
      \hline
      \textbf{MatchSum + BERT} & 43.98 & 20.50 & \cellcolor[gray]{0.9}28.62 \\
      \textbf{MatchSum + RoBERTa} & 44.17 & 20.80 & \cellcolor[gray]{0.9}28.86 \\
      
      \hline
      \textbf{CoLo + BERT} & 43.34 & 20.27 & \cellcolor[gray]{0.9}27.77 \\
      \textbf{CoLo} & 44.18 & 21.03 & \cellcolor[gray]{0.9}27.98 \\
      \textbf{CoLo 1024} & 44.41 & 21.19 & \cellcolor[gray]{0.9}27.98 \\
      
      \hline
      \textbf{OrderSum + BERT} & 43.44 & 20.39 & \cellcolor[gray]{0.9}29.32 \\
      \textbf{OrderSum} & 44.23 & 21.17 & \cellcolor[gray]{0.9}30.21 \\
      \textbf{OrderSum 1024} & \textbf{44.44} & \textbf{21.31} & \cellcolor[gray]{0.9}\textbf{30.52} \\
      
      \bottomrule 
    \end{tabular} 
  \normalsize
  \caption{Results on the CNN/DailyMail test set. $\textbf{R-1}$, $\textbf{R-2}$, $\textbf{R-L}$ refer to ROUGE-1, ROUGE-2, and ROUGE-L scores with reference summaries, respectively. BERTSUM-EXT \citep{liu-lapata-2019-text} is a variant of BERTSUM where the encoder is followed by two additional transformer layers. MatchSum + BERT and MatchSum + RoBERTa are MatchSum using BERT and RoBERTa \citep{liu_roberta_2019} as the reranker models, respectively \citep{zhong-etal-2020-extractive}. CoLo + BERT and OrderSum + BERT use BERTSUM, while CoLo and OrderSum use BARTSUM as initialization models. CoLo 1024 and OrderSum 1024 are initialized with BARTSUM 1024.}
  \label{table:3}
\end{table}

To verify the effectiveness of OrderSum in improving the sentence order of extractive summaries, we evaluate the models from the sentence-level framework, including BERTSUM \citep{liu-lapata-2019-text} and BARTSUM \citep{lewis-etal-2020-bart}, and the summary-level framework, including MatchSum \citep{zhong-etal-2020-extractive} and CoLo \citep{an-etal-2022-colo}, as the baselines. In particular, CoLo using BARTSUM with a sequence length of 1024 is the state-of-the-art model. CoLo has a similar model structure to our model, but does not consider sentence order across the entire process. For detailed analysis, we retrained CoLo with higher scores than those reported in the paper\footnote{We use the \textit{rouge-score} library, developed by Google Research. It calculates the ROUGE scores lower than the \textit{pyrouge} library. For example, the ROUGE-1 and ROUGE-2 scores of LEAD on CNN/DailyMail drop from 40.42 to 40.14 and 17.62 to 17.55, respectively.}. Finally, given the recent emergence of applying LLMs in extractive summarization \citep{zhang-etal-2023-extractive-summarization, ge-etal-2024-extractive-summarization}, we conclude the section by evaluating the performance of LLM on the four extractive summarization datasets.


\begin{table*}[t!]
  \centering
  \small
  \setlength{\tabcolsep}{8.6pt}
  \renewcommand{\arraystretch}{1.30} 
  \begin{tabular}{l*{9}{c}}
    \toprule
    \multicolumn{1}{l}{\textbf{}} & \multicolumn{3}{c}{\textbf{XSum}} & \multicolumn{3}{c}{\textbf{WikiHow}} & \multicolumn{3}{c}{\textbf{PubMed}} \\

    \cmidrule(lr){2-4} \cmidrule(lr){5-7} \cmidrule(lr){8-10}
    & \textbf{R-1} & \textbf{R-2} & \cellcolor[gray]{0.9}\textbf{R-L} & \textbf{R-1} & \textbf{R-2} & \cellcolor[gray]{0.9}\textbf{R-L} & \textbf{R-1} & \textbf{R-2} & \cellcolor[gray]{0.9}\textbf{R-L} \\

    \hline
    \textbf{LEAD} & 16.03 & 1.58 & \cellcolor[gray]{0.9}11.80 & 24.93 & 5.84 & \cellcolor[gray]{0.9}15.55 & 37.19 & 12.18 & \cellcolor[gray]{0.9}19.84 \\
    
    \textbf{ORACLE} & 31.14 & 9.08 & \cellcolor[gray]{0.9}20.54 & 39.73 & 14.92 & \cellcolor[gray]{0.9}25.43 & 45.55 & 20.22 & \cellcolor[gray]{0.9}25.41 \\

    \textbf{Ordered ORACLE} & 31.14 & 9.08 & \cellcolor[gray]{0.9}22.21 & 39.73 & 14.94 & \cellcolor[gray]{0.9}27.15 & 45.55 & 20.23 & \cellcolor[gray]{0.9}28.48 \\

    \hline
    \textbf{BARTSUM} & 24.12 & 5.10 & \cellcolor[gray]{0.9}16.12 & 30.47 & 8.78 & \cellcolor[gray]{0.9}17.99 & 41.04 & \textbf{15.04} & \cellcolor[gray]{0.9}20.67 \\

    \textbf{MatchSum + BERT} & 24.60 & 4.71 & \cellcolor[gray]{0.9}16.23 & 31.82 & 8.98 & \cellcolor[gray]{0.9}20.32 & 40.82 & 14.85 & \cellcolor[gray]{0.9}\textbf{21.38} \\
    
    \textbf{CoLo} & 24.80 & \textbf{5.39} & \cellcolor[gray]{0.9}16.90 & 32.38 & 9.47 & \cellcolor[gray]{0.9}19.59 & 41.37 & 14.45 & \cellcolor[gray]{0.9}20.53 \\

    \textbf{OrderSum} & \textbf{24.81} & 5.33 & \cellcolor[gray]{0.9}\textbf{17.49} & \textbf{32.98} & \textbf{9.69} & \cellcolor[gray]{0.9}\textbf{21.05} & \textbf{41.58} & 14.64 & \cellcolor[gray]{0.9}21.19 \\

  \bottomrule
  \end{tabular}
  \caption{Results on the XSum, WikiHow, PubMed test sets. CoLo and OrderSum are initialized with the same parameters of BARTSUM on each dataset. In WikiHow and PubMed, the results of OrderSum are obtained from the model trained using anchor candidate sampling with a sampling factor of 2 (see in Table 9 in Appendix 2).}
  \label{table:4}
\normalsize
\end{table*}

\subsection{Results on CNN/DailyMail}

Table 3 presents the experimental results on CNN/DailyMail. The first block shows the results of LEAD, ORACLE, and Ordered ORACLE, denoting the first three sentences, the existing oracle summary, and the oracle summary considering the sentence order, respectively. Because the evaluation calculates the ROUGE scores between extractive summaries and abstractive reference summaries, LEAD and Ordered ORACLE serve as the lower and upper bounds. As described in Section 4.2, Ordered ORACLE increases ROUGE-L by 4.13 compared to ORACLE, while maintaining ROUGE-1 and slightly increasing ROUGE-2.

The second block shows the results of the sentence-level framework models. Notably, BARTSUM significantly outperforms BERTSUM on all ROUGE scores. BARTSUM 1024 further improves performance by increasing the input sequence length from 512 to 1024.

The third and fourth blocks show the results of the summary-level framework models, MatchSum and CoLo, respectively. When using the same extractor, the summary-level framework models outperform the sentence-level framework models on ROUGE-1 and ROUGE-2, metrics for evaluating sentence inclusion. Comparing MatchSum and CoLo, with BERTSUM as the extractor, CoLo + BERT underperforms MatchSum. However, CoLo outperforms MatchSum on ROUGE-1 and ROUGE-2 using BARTSUM as the extractor. The important point is that the existing summary-level framework models do not achieve meaningful improvements or perform worse than the sentence-level framework models on ROUGE-L, meaning they do not improve sentence order over the sentence-level framework.

The last block presents the results of OrderSum. Compared to the sentence-level framework models, OrderSum significantly increases ROUGE-L by up to 2.18 (BARTSUM 1024 vs. OrderSum 1024), in contrast to the existing summary-level framework models. Compared to the summary-level framework models, OrderSum outperforms CoLo on ROUGE-L by up to 2.54 (CoLo 1024 vs. OrderSum 1024), and even outperforms MatchSum + RoBERTa on ROUGE-L by 1.66, which consists of two different models each for the extractor and the reranker. This improvement is significant, given that the ROUGE-L increases from LEAD to BARTSUM 1024, CoLo 1024, MatchSum + RoBERTa, and OrderSum 1024 are 3.26, 2.90, 3.78, and 5.44, respectively.


\subsection{Results on XSum, WikiHow and PubMed}

Table 4 shows the results on XSum, WikiHow, and PubMed. On XSum, OrderSum outperforms BARTSUM, MatchSum + BERT, and CoLo on ROUGE-L by 1.37, 1.26, and 0.59, respectively. On WikiHow, OrderSum outperforms BARTSUM, MatchSum + BERT, and CoLo on ROUGE-L by 3.06, 0.73, and 1.46, respectively. These improvements are significant, considering that the ROUGE-L scores of Ordered ORACLE in XSum (22.21) and WikiHow (27.15) are lower than 44.11 in CNN/DailyMail, which is an upper bound for evaluation. On the other hand, OrderSum is less effective in improving ROUGE-L on PubMed. While OrderSum outperforms BARTSUM and CoLo, it slightly underperforms compared to MatchSum + BERT by 0.19. This indicates that the performance improvement in sentence order using OrderSum is influenced by the number of sentences in a targeted extractive summary, denoted as $\mathcal{N}$ in Table 2.

Notably, OrderSum achieves the highest scores on ROUGE-1 across the four datasets. This result indicates that the new representation method for embedding candidate summaries and the objective function for training sentence order do not harm but rather improve the performance in determining which sentences to include in the summary.


\subsection{Validation Graphs for ROUGE-L during Training}

\begin{figure*}[t!]
  \centering
  \includegraphics[width=\linewidth]{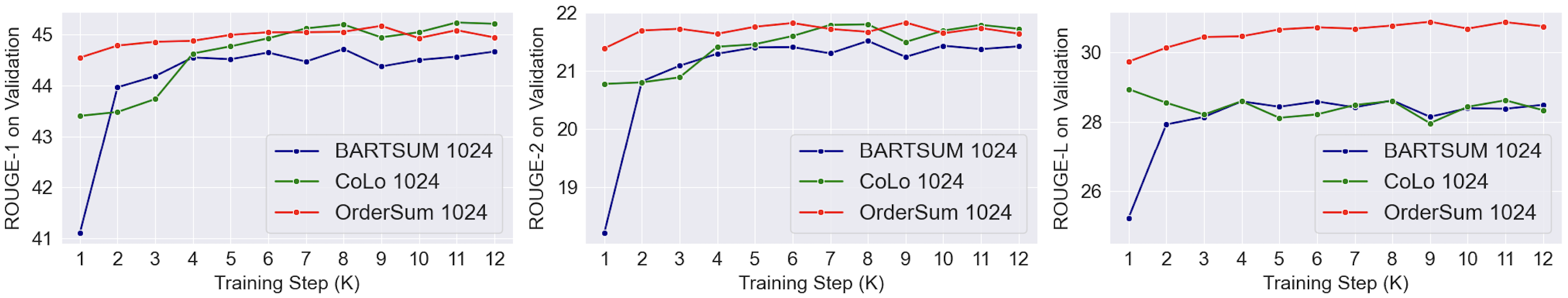}
  \caption{Validation graphs for ROUGE-1, ROUGE-2, and ROUGE-L scores during the training of BARTSUM 1024, CoLo 1024, and OrderSum 1024 on CNN/DailyMail. Training is conducted for 12K steps, with validation performed every 1,000 steps.}
  \label{fig:fig2}
\end{figure*}

\begin{figure*}[t!]
  \centering
  \includegraphics[width=\linewidth]{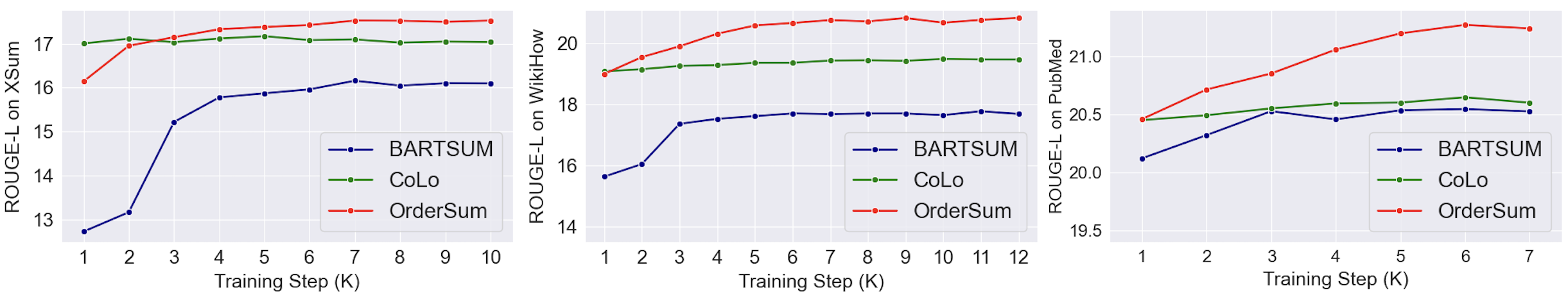}
  \caption{Validation graphs for ROUGE-L scores during the training of BARTSUM, CoLo, and OrderSum on the three datasets. Training is conducted for 10K, 12K, and 7K steps on XSum, WikiHow, and PubMed, respectively, with validation performed every 1,000 steps.}
  \label{fig:fig3}
\end{figure*}

To verify that OrderSum trains the sentence order, we present validation graphs based on models trained on the four datasets. Figure 2 shows the validation graphs for ROUGE-1, ROUGE-2, and ROUGE-L scores obtained during the training of BARTSUM 1024, CoLo 1024, and OrderSum 1024 on CNN/DailyMail. In the ROUGE-L plot, the OrderSum 1024 graph increases continuously, ending with a significant gap compared to the graphs of the other models. In contrast, the CoLo 1024 graph in the ROUGE-L plot does not show a meaningful increase from the beginning of training, supporting the results in Table 3. This indicates that OrderSum 1024 trains sentence order using a new objective function that includes ROUGE-L, unlike the previous models.

Figure 3 presents the validation graphs for ROUGE-L during the training of the three models, BARTSUM, CoLo, and OrderSum, on XSum, WikiHow, and PubMed. The graphs for OrderSum show an increase across all three datasets, indicating that OrderSum trains the sentence order on different datasets. In particular, in the case of PubMed, the OrderSum graph increases continuously, although the gain is not large. The validation graphs for all ROUGE scores on the four datasets are shown in Appendix 3.


\subsection{Qualitative Analysis}

\begin{table*}[t!]
  \centering
  \setlength{\tabcolsep}{12pt} 
  \renewcommand{\arraystretch}{1.3} 
  \small
    \begin{tabular}{lp{12cm}}
    \toprule
    \multicolumn{1}{c}{\small \textbf{System}} & \multicolumn{1}{c}{\small \textbf{Summary}} \\
    
    \hline
    \textbf{Document} & Passengers of an Icelandair flight to Denver on Tuesday were left in a bit of shock after their plane was struck by lightning shortly after takeoff. Though the Boeing 757 continued its 3,700-mile journey from Reykjavik, Iceland, to Denver, it wasn't until the plane landed that passengers and crew were aware of a gaping hole in the nose of the plane. The hole was at a point in the plane where weather radars are housed, ....  \\
    
    \hline
    \textbf{Reference} & Flight was traveling from Reykjavik, Iceland to Denver when it was struck. Passengers said it was hit by lightning shortly after the plane took off. Pilots reported the lighting and continued eight-hour flight to Denver. It wasn’t until they landed that pilots notice huge hole at the nose of plane. No one on board was injured and the plane landed safely in Denver. \\
    
    \hline
    \textbf{CoLo 1024} & The hole was at a point in the plane where weather radars are housed, but the plane landed safely in Denver and no one was injured. A plane was struck by lightning shortly after takeoff during a flight from Reykjavik, Iceland, to Denver, Colorado on Tuesday. Though the Boeing 757 continued its 3,700-mile journey from Reykjavik, Iceland, to Denver, it wasn’t until the plane landed that passengers and crew were aware of a gaping hole in the nose of the plane. \\
    
    \hline
    \textbf{OrderSum 1024} & A plane was struck by lightning shortly after takeoff during a flight from Reykjavik, Iceland, to Denver, Colorado on Tuesday. It wasn’t until after landing that the passengers and crew found out the lightning strike caused a gaping hole at the nose of the plane. The hole was at a point in the plane where weather radars are housed, but the plane landed safely in Denver and no one was injured. \\
    
    \bottomrule 
    \end{tabular} 
  \normalsize
  \caption{Example from the CNN/DailyMail test set. The document is a news report about a plane that landed safely after being struck by lightning during a flight. The reference summary is a human-written abstractive summary.}
  \label{table:5}
\end{table*}

Table 5 compares the summaries generated by OrderSum 1024 and CoLo 1024 to demonstrate how OrderSum qualitatively improves the sentence order of extractive summaries. First, the summary of OrderSum 1024 has a natural flow, with its sentence order aligned with the abstractive reference summary. The first sentence of the OrderSum 1024 summary corresponds to the first two sentences of the reference summary, which introduce the incident. The second and third sentences correspond to the fourth and fifth sentences of the reference summary, respectively. On the other hand, the sentence order of the CoLo 1024 summary does not align with the reference summary. The first sentence of the CoLo 1024 summary corresponds to the last sentence of the reference summary, the second sentence corresponds to the beginning of the reference summary, and the final sentence corresponds to the fourth sentence of the reference summary. Although the CoLo 1024 summary includes all the necessary sentences, its quality is poor because the sentences are not properly ordered. See Appendix 5 for more examples from the four datasets.


\subsection{Comparison of Sentence Order: Sentence-Level vs. Summary-Level}

\begin{table}[t!]
  \centering
  \small
  \setlength{\tabcolsep}{4.5pt}
  \renewcommand{\arraystretch}{1.05} 
  \begin{tabular}{lcccc}
    \toprule
    \textbf{} & \textbf{CNN/DM} & \textbf{XSum} & \textbf{WikiHow} & \textbf{PubMed} \\

    \midrule
    \textbf{R-L} & 30.52 & 17.49 & 21.05 & 21.19 \\
    \textbf{$\textbf{R-L}_{ext}$} & 28.72 & 17.30 & 19.93 & 20.59 \\
  \bottomrule
  \end{tabular}
  \caption{$\textbf{R-L}$ denotes the results of OrderSum, and $\textbf{R-L}_{ext}$ denotes the results of the summaries reordered using $p_{f_{\theta}}$ obtained by the extractor. The two summaries for each data share the same sentences but differ in sentence order. OrderSum 1024 is used for CNN/DailyMail, and OrderSum is used for the remaining datasets.}
  \label{table:6}
\normalsize
\end{table}

\begin{table}[t!]
  \centering
  \small
  \setlength{\tabcolsep}{6.8pt}
  \renewcommand{\arraystretch}{0.92} 
  \begin{tabular}{lcccc}
    \toprule
    \textbf{} & \textbf{CNN/DM} & \textbf{XSum} & \textbf{WikiHow} & \textbf{PubMed} \\

    \midrule
    \textbf{$\rho$} & 0.297 & 0.680 & 0.117 & 0.196 \\
  \bottomrule
  \end{tabular}
  \caption{The Spearman's rank correlation coefficients between the OrderSum summaries and the summaries reordered by $p_{f_{\theta}}$. Spearman's rank correlation calculates the Pearson correlation between the ranks of the variables instead of the variables. Spearman's rank correlation coefficient ranges from -1 to 1. The closer to 1, the stronger the correlation between the ranks.}
  \label{table:7}
\normalsize
\end{table}

To demonstrate the importance of ordering sentences at the summary level, we generate extractive summaries in which the sentences are reordered based on the probabilities obtained by the extractor, $p_{f_{\theta}}$, from the results of OrderSum. Since the extractor is a sentence-level framework model, the sentence order sorted by $p_{f_{\theta}}$ mimics the sentence order determined at the sentence level.

In Table 6, $\textbf{R-L}$ is the results of OrderSum, and $\textbf{R-L}_{ext}$ is the results of the summaries reordered by $p_{f_{\theta}}$. On CNN/DailyMail, WikiHow, and PubMed, the difference between $\textbf{R-L}$ and $\textbf{R-L}_{ext}$ is 1.8, 1.12, and 0.6, respectively. These results show that ordering sentences at the summary level by reranking candidate summaries is critical for improving sentence order. In contrast, on XSum, the decrease from $\textbf{R-L}$ to $\textbf{R-L}_{ext}$ is only 0.19. However, $\textbf{R-L}_{ext}$ increases by 0.40 compared to the result of CoLo on XSum, 16.90, shown in Table 4. This suggests that extracting key sentences contributes more than reranking candidate summaries on XSum, as the parameters of the extractor are also updated by the new objective function, as mentioned in Section 4.3.

Table 7 shows the statistical difference in sentence order between the OrderSum summaries and the summaries reordered by $p_{f_{\theta}}$, using Spearman's rank correlation. In the table, $\rho$ denotes the Spearman's rank correlation coefficient between the two summaries. On CNN/DailyMail, WikiHow, and PubMed, the coefficients are 0.297, 0.117, and 0.196, respectively, indicating weak correlations between the two summaries. This means that the sentence order determined by the extractor is dramatically changed by reranking candidate summaries. On the other hand, the coefficient on XSum is 0.680, which is relatively close to 1, indicating that reranking candidate summaries does not significantly affect the sentence order.

The consistent results from the two tables demonstrate that ordering sentences at the summary level plays a crucial role in improving the sentence order of extractive summaries.


\subsection{LLM for Extractive Summarization}

\begin{table}[h!]
  \centering
  \setlength{\tabcolsep}{13pt} 
  \renewcommand{\arraystretch}{1.18} 
  \small
    \begin{tabular}{lccc}
      \toprule
      \multicolumn{1}{c}{\textbf{}} & \textbf{R-1} & \textbf{R-2} & \textbf{R-L}\\
      
      \midrule
      \textbf{CNN/DM} & 38.16 & 15.77 & 24.37 \\
      \textbf{XSum} & 19.98 & 3.15 & 13.23 \\
      \textbf{WikiHow} & 27.04 & 6.99 & 17.08 \\
      \textbf{OrderSum} & 39.47 & 12.83 & 19.41 \\
      
      \bottomrule 
    \end{tabular} 
  \normalsize
  \caption{We use the "gpt-4o-2024-08-06" version of GPT-4o as a test model, one of the LLMs with the highest generalization capabilities.}
  \label{table:8}
\end{table}

Finally, we present the results of GPT-4o performing extractive summarization on the four datasets in Table 8. To provide instruction about the extractive summarization task, we include three-shot examples consisting of documents, reference summaries, and Ordered ORACLEs in the prompt \citep{zhang-etal-2023-extractive-summarization} (see the prompt in Appendix 4).

On CNN/DailyMail, GPT-4o has slightly lower scores than LEAD in Table 3, generated by selecting the first three sentences of the document. On XSum, WikiHow, and PubMed, GPT-4o outperforms LEAD but still underperforms the existing extractive summarization models. We could not obtain results comparable to those reported in \citet{zhang-etal-2023-extractive-summarization}, which evaluated the model using only 1,000 samples from each dataset's test set, despite the similar instruction setting. These results show that GPT-4o, despite its strong generalization capabilities, does not reach the performance of the existing models without extra training for the extractive summarization task. We leave the experiments on post-training of the LLM for the extractive summarization as future work.


\section{Conclusion}

We propose OrderSum, a novel extractive summarization model that semantically orders sentences within an extractive summary. OrderSum represents embeddings of extractive summaries to incorporate the sentence order and trains the model to identify which extractive summary has a better sentence order in the semantic space. As a result, OrderSum achieves state-of-the-art performance on ROUGE-L, the metric for evaluating sentence order, significantly outperforming existing models. Additional qualitative and statistical analyses demonstrate how OrderSum improves the sentence order of extractive summaries compared to the existing methods.

The limitation of our study is that performance improvement becomes less effective when the number of sentences in the extractive summary increases excessively, exceeding five in our experiments. We suggest the following reasons: 1) using permutation to generate candidate summaries significantly increases the number of candidates and leads to the degradation of contrastive learning. 2) As the number of sentences in the summary increases, the embedding of each sentence within the fixed-size candidate summary embedding becomes more compressed. We will leave these limitations for future work. Lastly, we plan to introduce the objective function of OrderSum to abstractive summarization.


\nocite{20,
liu-lapata-2019-text,
zhong-etal-2020-extractive,
an-etal-2022-colo,
lewis-etal-2020-bart,
NIPS2015_afdec700,
narayan-etal-2018-dont,
koupaee2018wikihow,
cohan-etal-2018-discourse,
cheng-lapata-2016-neural,
nallapati_summarunner_2017,
zhang-etal-2018-neural,
zhou-etal-2018-neural-document,
narayan-etal-2018-ranking,
bae-etal-2019-summary,
dong-etal-2018-banditsum,
zhang-etal-2023-diffusum,
mihalcea-tarau-2004-textrank,
erkan_lexrank_2004,
lin-2004-rouge,
Zhang*2020BERTScore:,
NEURIPS2021_e4d2b6e6,
zhao-etal-2019-moverscore,
ijcai2019p0748,
10.5555/3504035.3504683,
oord2019representationlearningcontrastivepredictive,
Schroff_2015_CVPR,
lin-hovy-2003-potential,
hirao-etal-2017-enumeration,
NIPS2017_3f5ee243,
loshchilov2018decoupled,
liu_roberta_2019,
radford2019language,
NEURIPS2020_1457c0d6,
zhang-etal-2023-extractive-summarization,
ge-etal-2024-extractive-summarization,
rush-etal-2015-neural,
nallapati-etal-2016-abstractive,
liu-liu-2021-simcls,
devlin-etal-2019-bert,
radev2002evaluation,
}

\bibliography{reference}

\begin{thebibliography}{40}
\expandafter\ifx\csname natexlab\endcsname\relax\def\natexlab#1{#1}\fi

\bibitem[{An et~al.(2022)An, Zhong, Wu, Zhu, Huang, and Qiu}]{an-etal-2022-colo}
Chenxin An, Ming Zhong, Zhiyong Wu, Qin Zhu, Xuanjing Huang, and Xipeng Qiu. 2022.
\newblock \href {https://aclanthology.org/2022.coling-1.508/} {{C}o{L}o: A contrastive learning based re-ranking framework for one-stage summarization}.
\newblock In \emph{Proceedings of the 29th International Conference on Computational Linguistics}, pages 5783--5793, Gyeongju, Republic of Korea. International Committee on Computational Linguistics.

\bibitem[{Bae et~al.(2019)Bae, Kim, Kim, and Lee}]{bae-etal-2019-summary}
Sanghwan Bae, Taeuk Kim, Jihoon Kim, and Sang-goo Lee. 2019.
\newblock \href {https://doi.org/10.18653/v1/D19-5402} {Summary level training of sentence rewriting for abstractive summarization}.
\newblock In \emph{Proceedings of the 2nd Workshop on New Frontiers in Summarization}, pages 10--20, Hong Kong, China. Association for Computational Linguistics.

\bibitem[{Brown et~al.(2020)Brown, Mann, Ryder, Subbiah, Kaplan, Dhariwal, Neelakantan, Shyam, Sastry, Askell, Agarwal, Herbert-Voss, Krueger, Henighan, Child, Ramesh, Ziegler, Wu, Winter, Hesse, Chen, Sigler, Litwin, Gray, Chess, Clark, Berner, McCandlish, Radford, Sutskever, and Amodei}]{NEURIPS2020_1457c0d6}
Tom Brown, Benjamin Mann, Nick Ryder, Melanie Subbiah, Jared~D Kaplan, Prafulla Dhariwal, Arvind Neelakantan, Pranav Shyam, Girish Sastry, Amanda Askell, Sandhini Agarwal, Ariel Herbert-Voss, Gretchen Krueger, Tom Henighan, Rewon Child, Aditya Ramesh, Daniel Ziegler, Jeffrey Wu, Clemens Winter, Chris Hesse, Mark Chen, Eric Sigler, Mateusz Litwin, Scott Gray, Benjamin Chess, Jack Clark, Christopher Berner, Sam McCandlish, Alec Radford, Ilya Sutskever, and Dario Amodei. 2020.
\newblock \href {https://proceedings.neurips.cc/paper_files/paper/2020/file/1457c0d6bfcb4967418bfb8ac142f64a-Paper.pdf} {Language models are few-shot learners}.
\newblock In \emph{Advances in Neural Information Processing Systems}, volume~33, pages 1877--1901. Curran Associates, Inc.

\bibitem[{Cheng and Lapata(2016)}]{cheng-lapata-2016-neural}
Jianpeng Cheng and Mirella Lapata. 2016.
\newblock \href {https://doi.org/10.18653/v1/P16-1046} {Neural summarization by extracting sentences and words}.
\newblock In \emph{Proceedings of the 54th Annual Meeting of the Association for Computational Linguistics (Volume 1: Long Papers)}, pages 484--494, Berlin, Germany. Association for Computational Linguistics.

\bibitem[{Cohan et~al.(2018)Cohan, Dernoncourt, Kim, Bui, Kim, Chang, and Goharian}]{cohan-etal-2018-discourse}
Arman Cohan, Franck Dernoncourt, Doo~Soon Kim, Trung Bui, Seokhwan Kim, Walter Chang, and Nazli Goharian. 2018.
\newblock \href {https://doi.org/10.18653/v1/N18-2097} {A discourse-aware attention model for abstractive summarization of long documents}.
\newblock In \emph{Proceedings of the 2018 Conference of the North {A}merican Chapter of the Association for Computational Linguistics: Human Language Technologies, Volume 2 (Short Papers)}, pages 615--621, New Orleans, Louisiana. Association for Computational Linguistics.

\bibitem[{Devlin et~al.(2019)Devlin, Chang, Lee, and Toutanova}]{devlin-etal-2019-bert}
Jacob Devlin, Ming-Wei Chang, Kenton Lee, and Kristina Toutanova. 2019.
\newblock \href {https://doi.org/10.18653/v1/N19-1423} {{BERT}: Pre-training of deep bidirectional transformers for language understanding}.
\newblock In \emph{Proceedings of the 2019 Conference of the North {A}merican Chapter of the Association for Computational Linguistics: Human Language Technologies, Volume 1 (Long and Short Papers)}, pages 4171--4186, Minneapolis, Minnesota. Association for Computational Linguistics.

\bibitem[{Dong et~al.(2018)Dong, Shen, Crawford, van Hoof, and Cheung}]{dong-etal-2018-banditsum}
Yue Dong, Yikang Shen, Eric Crawford, Herke van Hoof, and Jackie Chi~Kit Cheung. 2018.
\newblock \href {https://doi.org/10.18653/v1/D18-1409} {{B}andit{S}um: Extractive summarization as a contextual bandit}.
\newblock In \emph{Proceedings of the 2018 Conference on Empirical Methods in Natural Language Processing}, pages 3739--3748, Brussels, Belgium. Association for Computational Linguistics.

\bibitem[{Erkan and Radev(2004)}]{erkan_lexrank_2004}
Günes Erkan and Dragomir~R. Radev. 2004.
\newblock Lexrank: {Graph}-based lexical centrality as salience in text summarization.
\newblock \emph{Journal of artificial intelligence research}, 22:457--479.

\bibitem[{Ge et~al.(2024)Ge, Jeoung, and Diesner}]{ge-etal-2024-extractive-summarization}
Yubin Ge, Sullam Jeoung, and Jana Diesner. 2024.
\newblock \href {https://aclanthology.org/2024.inlg-main.10/} {Extractive summarization via fine-grained semantic tuple extraction}.
\newblock In \emph{Proceedings of the 17th International Natural Language Generation Conference}, pages 121--133, Tokyo, Japan. Association for Computational Linguistics.

\bibitem[{Hermann et~al.(2015)Hermann, Kocisky, Grefenstette, Espeholt, Kay, Suleyman, and Blunsom}]{NIPS2015_afdec700}
Karl~Moritz Hermann, Tomas Kocisky, Edward Grefenstette, Lasse Espeholt, Will Kay, Mustafa Suleyman, and Phil Blunsom. 2015.
\newblock \href {https://proceedings.neurips.cc/paper_files/paper/2015/file/afdec7005cc9f14302cd0474fd0f3c96-Paper.pdf} {Teaching machines to read and comprehend}.
\newblock In \emph{Advances in Neural Information Processing Systems}, volume~28. Curran Associates, Inc.

\bibitem[{Hirao et~al.(2017)Hirao, Nishino, Suzuki, and Nagata}]{hirao-etal-2017-enumeration}
Tsutomu Hirao, Masaaki Nishino, Jun Suzuki, and Masaaki Nagata. 2017.
\newblock \href {https://aclanthology.org/E17-1037/} {Enumeration of extractive oracle summaries}.
\newblock In \emph{Proceedings of the 15th Conference of the {E}uropean Chapter of the Association for Computational Linguistics: Volume 1, Long Papers}, pages 386--396, Valencia, Spain. Association for Computational Linguistics.

\bibitem[{Koupaee and Wang(2018)}]{koupaee2018wikihow}
Mahnaz Koupaee and William~Yang Wang. 2018.
\newblock \href {http://arxiv.org/abs/1810.09305} {Wikihow: A large scale text summarization dataset}.

\bibitem[{Lewis et~al.(2020)Lewis, Liu, Goyal, Ghazvininejad, Mohamed, Levy, Stoyanov, and Zettlemoyer}]{lewis-etal-2020-bart}
Mike Lewis, Yinhan Liu, Naman Goyal, Marjan Ghazvininejad, Abdelrahman Mohamed, Omer Levy, Veselin Stoyanov, and Luke Zettlemoyer. 2020.
\newblock \href {https://doi.org/10.18653/v1/2020.acl-main.703} {{BART}: Denoising sequence-to-sequence pre-training for natural language generation, translation, and comprehension}.
\newblock In \emph{Proceedings of the 58th Annual Meeting of the Association for Computational Linguistics}, pages 7871--7880, Online. Association for Computational Linguistics.

\bibitem[{Lin(2004)}]{lin-2004-rouge}
Chin-Yew Lin. 2004.
\newblock \href {https://aclanthology.org/W04-1013/} {{ROUGE}: A package for automatic evaluation of summaries}.
\newblock In \emph{Text Summarization Branches Out}, pages 74--81, Barcelona, Spain. Association for Computational Linguistics.

\bibitem[{Lin and Hovy(2003)}]{lin-hovy-2003-potential}
Chin-Yew Lin and Eduard Hovy. 2003.
\newblock \href {https://aclanthology.org/W03-0510/} {The potential and limitations of automatic sentence extraction for summarization}.
\newblock In \emph{Proceedings of the {HLT}-{NAACL} 03 Text Summarization Workshop}, pages 73--80.

\bibitem[{Liu and Lapata(2019)}]{liu-lapata-2019-text}
Yang Liu and Mirella Lapata. 2019.
\newblock \href {https://doi.org/10.18653/v1/D19-1387} {Text summarization with pretrained encoders}.
\newblock In \emph{Proceedings of the 2019 Conference on Empirical Methods in Natural Language Processing and the 9th International Joint Conference on Natural Language Processing (EMNLP-IJCNLP)}, pages 3730--3740, Hong Kong, China. Association for Computational Linguistics.

\bibitem[{Liu et~al.(2019)Liu, Ott, Goyal, Du, Joshi, Chen, Levy, Lewis, Zettlemoyer, and Stoyanov}]{liu_roberta_2019}
Yinhan Liu, Myle Ott, Naman Goyal, Jingfei Du, Mandar Joshi, Danqi Chen, Omer Levy, Mike Lewis, Luke Zettlemoyer, and Veselin Stoyanov. 2019.
\newblock {RoBERTa}: {A} {Robustly} {Optimized} {BERT} {Pretraining} {Approach}.
\newblock ArXiv:1907.11692 [cs].

\bibitem[{Liu and Liu(2021)}]{liu-liu-2021-simcls}
Yixin Liu and Pengfei Liu. 2021.
\newblock \href {https://doi.org/10.18653/v1/2021.acl-short.135} {{S}im{CLS}: A simple framework for contrastive learning of abstractive summarization}.
\newblock In \emph{Proceedings of the 59th Annual Meeting of the Association for Computational Linguistics and the 11th International Joint Conference on Natural Language Processing (Volume 2: Short Papers)}, pages 1065--1072, Online. Association for Computational Linguistics.

\bibitem[{Logeswaran et~al.(2018)Logeswaran, Lee, and Radev}]{10.5555/3504035.3504683}
Lajanugen Logeswaran, Honglak Lee, and Dragomir Radev. 2018.
\newblock Sentence ordering and coherence modeling using recurrent neural networks.
\newblock In \emph{Proceedings of the Thirty-Second AAAI Conference on Artificial Intelligence and Thirtieth Innovative Applications of Artificial Intelligence Conference and Eighth AAAI Symposium on Educational Advances in Artificial Intelligence}, AAAI'18/IAAI'18/EAAI'18. AAAI Press.

\bibitem[{Loshchilov and Hutter(2019)}]{loshchilov2018decoupled}
Ilya Loshchilov and Frank Hutter. 2019.
\newblock \href {https://openreview.net/forum?id=Bkg6RiCqY7} {Decoupled weight decay regularization}.
\newblock In \emph{International Conference on Learning Representations}.

\bibitem[{Mihalcea and Tarau(2004)}]{mihalcea-tarau-2004-textrank}
Rada Mihalcea and Paul Tarau. 2004.
\newblock \href {https://aclanthology.org/W04-3252/} {{T}ext{R}ank: Bringing order into text}.
\newblock In \emph{Proceedings of the 2004 Conference on Empirical Methods in Natural Language Processing}, pages 404--411, Barcelona, Spain. Association for Computational Linguistics.

\bibitem[{Nallapati et~al.(2017)Nallapati, Zhai, and Zhou}]{nallapati_summarunner_2017}
Ramesh Nallapati, Feifei Zhai, and Bowen Zhou. 2017.
\newblock Summarunner: {A} recurrent neural network based sequence model for extractive summarization of documents.
\newblock In \emph{Proceedings of the {AAAI} conference on artificial intelligence}, volume~31.
\newblock Issue: 1.

\bibitem[{Nallapati et~al.(2016)Nallapati, Zhou, dos Santos, Gu{\ensuremath{\dot{}}}l{\c{c}}ehre, and Xiang}]{nallapati-etal-2016-abstractive}
Ramesh Nallapati, Bowen Zhou, Cicero dos Santos, {\c{C}}a{\u{g}}lar Gu{\ensuremath{\dot{}}}l{\c{c}}ehre, and Bing Xiang. 2016.
\newblock \href {https://doi.org/10.18653/v1/K16-1028} {Abstractive text summarization using sequence-to-sequence {RNN}s and beyond}.
\newblock In \emph{Proceedings of the 20th {SIGNLL} Conference on Computational Natural Language Learning}, pages 280--290, Berlin, Germany. Association for Computational Linguistics.

\bibitem[{Narayan et~al.(2018{\natexlab{a}})Narayan, Cohen, and Lapata}]{narayan-etal-2018-dont}
Shashi Narayan, Shay~B. Cohen, and Mirella Lapata. 2018{\natexlab{a}}.
\newblock \href {https://doi.org/10.18653/v1/D18-1206} {Don`t give me the details, just the summary! topic-aware convolutional neural networks for extreme summarization}.
\newblock In \emph{Proceedings of the 2018 Conference on Empirical Methods in Natural Language Processing}, pages 1797--1807, Brussels, Belgium. Association for Computational Linguistics.

\bibitem[{Narayan et~al.(2018{\natexlab{b}})Narayan, Cohen, and Lapata}]{narayan-etal-2018-ranking}
Shashi Narayan, Shay~B. Cohen, and Mirella Lapata. 2018{\natexlab{b}}.
\newblock \href {https://doi.org/10.18653/v1/N18-1158} {Ranking sentences for extractive summarization with reinforcement learning}.
\newblock In \emph{Proceedings of the 2018 Conference of the North {A}merican Chapter of the Association for Computational Linguistics: Human Language Technologies, Volume 1 (Long Papers)}, pages 1747--1759, New Orleans, Louisiana. Association for Computational Linguistics.

\bibitem[{van~den Oord et~al.(2019)van~den Oord, Li, and Vinyals}]{oord2019representationlearningcontrastivepredictive}
Aaron van~den Oord, Yazhe Li, and Oriol Vinyals. 2019.
\newblock \href {http://arxiv.org/abs/1807.03748} {Representation learning with contrastive predictive coding}.

\bibitem[{Radev et~al.(2002)Radev, Teufel, Saggion, Lam, Blitzer, Celebi, Qi, Drabek, and Liu}]{radev2002evaluation}
Dragomir Radev, Simone Teufel, Horacio Saggion, Wai Lam, John Blitzer, Arda Celebi, Hong Qi, Elliott Drabek, and Danyu Liu. 2002.
\newblock Evaluation of text summarization in a cross-lingual information retrieval framework.
\newblock \emph{Center for Language and Speech Processing, Johns Hopkins University, Baltimore, MD, Tech. Rep}, 6.

\bibitem[{Radford et~al.(2019)Radford, Wu, Child, Luan, Amodei, and Sutskever}]{radford2019language}
Alec Radford, Jeff Wu, Rewon Child, David Luan, Dario Amodei, and Ilya Sutskever. 2019.
\newblock Language models are unsupervised multitask learners.

\bibitem[{Rush et~al.(2015)Rush, Chopra, and Weston}]{rush-etal-2015-neural}
Alexander~M. Rush, Sumit Chopra, and Jason Weston. 2015.
\newblock \href {https://doi.org/10.18653/v1/D15-1044} {A neural attention model for abstractive sentence summarization}.
\newblock In \emph{Proceedings of the 2015 Conference on Empirical Methods in Natural Language Processing}, pages 379--389, Lisbon, Portugal. Association for Computational Linguistics.

\bibitem[{Schroff et~al.(2015)Schroff, Kalenichenko, and Philbin}]{Schroff_2015_CVPR}
Florian Schroff, Dmitry Kalenichenko, and James Philbin. 2015.
\newblock Facenet: A unified embedding for face recognition and clustering.
\newblock In \emph{Proceedings of the IEEE Conference on Computer Vision and Pattern Recognition (CVPR)}.

\bibitem[{Vaswani et~al.(2017)Vaswani, Shazeer, Parmar, Uszkoreit, Jones, Gomez, Kaiser, and Polosukhin}]{NIPS2017_3f5ee243}
Ashish Vaswani, Noam Shazeer, Niki Parmar, Jakob Uszkoreit, Llion Jones, Aidan~N Gomez, \L~ukasz Kaiser, and Illia Polosukhin. 2017.
\newblock \href {https://proceedings.neurips.cc/paper_files/paper/2017/file/3f5ee243547dee91fbd053c1c4a845aa-Paper.pdf} {Attention is all you need}.
\newblock In \emph{Advances in Neural Information Processing Systems}, volume~30. Curran Associates, Inc.

\bibitem[{Yin et~al.(2019)Yin, Song, Su, Zeng, Zhou, and Luo}]{ijcai2019p0748}
Yongjing Yin, Linfeng Song, Jinsong Su, Jiali Zeng, Chulun Zhou, and Jiebo Luo. 2019.
\newblock \href {https://doi.org/10.24963/ijcai.2019/748} {Graph-based neural sentence ordering}.
\newblock In \emph{Proceedings of the Twenty-Eighth International Joint Conference on Artificial Intelligence, {IJCAI-19}}, pages 5387--5393. International Joint Conferences on Artificial Intelligence Organization.

\bibitem[{Yuan et~al.(2021)Yuan, Neubig, and Liu}]{NEURIPS2021_e4d2b6e6}
Weizhe Yuan, Graham Neubig, and Pengfei Liu. 2021.
\newblock \href {https://proceedings.neurips.cc/paper_files/paper/2021/file/e4d2b6e6fdeca3e60e0f1a62fee3d9dd-Paper.pdf} {Bartscore: Evaluating generated text as text generation}.
\newblock In \emph{Advances in Neural Information Processing Systems}, volume~34, pages 27263--27277. Curran Associates, Inc.

\bibitem[{Zhang et~al.(2023{\natexlab{a}})Zhang, Liu, and Zhang}]{zhang-etal-2023-diffusum}
Haopeng Zhang, Xiao Liu, and Jiawei Zhang. 2023{\natexlab{a}}.
\newblock \href {https://doi.org/10.18653/v1/2023.findings-acl.828} {{D}iffu{S}um: Generation enhanced extractive summarization with diffusion}.
\newblock In \emph{Findings of the Association for Computational Linguistics: ACL 2023}, pages 13089--13100, Toronto, Canada. Association for Computational Linguistics.

\bibitem[{Zhang et~al.(2023{\natexlab{b}})Zhang, Liu, and Zhang}]{zhang-etal-2023-extractive-summarization}
Haopeng Zhang, Xiao Liu, and Jiawei Zhang. 2023{\natexlab{b}}.
\newblock \href {https://doi.org/10.18653/v1/2023.findings-emnlp.214} {Extractive summarization via {C}hat{GPT} for faithful summary generation}.
\newblock In \emph{Findings of the Association for Computational Linguistics: EMNLP 2023}, pages 3270--3278, Singapore. Association for Computational Linguistics.

\bibitem[{Zhang* et~al.(2020)Zhang*, Kishore*, Wu*, Weinberger, and Artzi}]{Zhang*2020BERTScore:}
Tianyi Zhang*, Varsha Kishore*, Felix Wu*, Kilian~Q. Weinberger, and Yoav Artzi. 2020.
\newblock \href {https://openreview.net/forum?id=SkeHuCVFDr} {Bertscore: Evaluating text generation with bert}.
\newblock In \emph{International Conference on Learning Representations}.

\bibitem[{Zhang et~al.(2018)Zhang, Lapata, Wei, and Zhou}]{zhang-etal-2018-neural}
Xingxing Zhang, Mirella Lapata, Furu Wei, and Ming Zhou. 2018.
\newblock \href {https://doi.org/10.18653/v1/D18-1088} {Neural latent extractive document summarization}.
\newblock In \emph{Proceedings of the 2018 Conference on Empirical Methods in Natural Language Processing}, pages 779--784, Brussels, Belgium. Association for Computational Linguistics.

\bibitem[{Zhao et~al.(2019)Zhao, Peyrard, Liu, Gao, Meyer, and Eger}]{zhao-etal-2019-moverscore}
Wei Zhao, Maxime Peyrard, Fei Liu, Yang Gao, Christian~M. Meyer, and Steffen Eger. 2019.
\newblock \href {https://doi.org/10.18653/v1/D19-1053} {{M}over{S}core: Text generation evaluating with contextualized embeddings and earth mover distance}.
\newblock In \emph{Proceedings of the 2019 Conference on Empirical Methods in Natural Language Processing and the 9th International Joint Conference on Natural Language Processing (EMNLP-IJCNLP)}, pages 563--578, Hong Kong, China. Association for Computational Linguistics.

\bibitem[{Zhong et~al.(2020)Zhong, Liu, Chen, Wang, Qiu, and Huang}]{zhong-etal-2020-extractive}
Ming Zhong, Pengfei Liu, Yiran Chen, Danqing Wang, Xipeng Qiu, and Xuanjing Huang. 2020.
\newblock \href {https://doi.org/10.18653/v1/2020.acl-main.552} {Extractive summarization as text matching}.
\newblock In \emph{Proceedings of the 58th Annual Meeting of the Association for Computational Linguistics}, pages 6197--6208, Online. Association for Computational Linguistics.

\bibitem[{Zhou et~al.(2018)Zhou, Yang, Wei, Huang, Zhou, and Zhao}]{zhou-etal-2018-neural-document}
Qingyu Zhou, Nan Yang, Furu Wei, Shaohan Huang, Ming Zhou, and Tiejun Zhao. 2018.
\newblock \href {https://doi.org/10.18653/v1/P18-1061} {Neural document summarization by jointly learning to score and select sentences}.
\newblock In \emph{Proceedings of the 56th Annual Meeting of the Association for Computational Linguistics (Volume 1: Long Papers)}, pages 654--663, Melbourne, Australia. Association for Computational Linguistics.

\end{thebibliography}
\bibliographystyle{acl_natbib}


\clearpage
\newpage
\appendix

\section*{Appendix 1}

ROUGE \citep{lin-2004-rouge} is a metric that evaluates a model-predicted text against a reference text by counting overlapping n-grams or word sequences between them. ROUGE-1 and ROUGE-2 measure the overlap of unigrams and bigrams obtained from the two texts. On the other hand, ROUGE-L finds the longest common subsequence (LCS) among word sequences that commonly appear in the same order in both texts. Because the words in the LCS do not need to be consecutive in the original text, ROUGE-L is primarily used to evaluate the structural similarity between the two texts.

There are two methods for evaluating summarization using ROUGE-L, depending on how the LCS between the reference summary $S^* = \{{s_1}^*, \ldots, {s_n}^*\}$ and the model summary $S = \{s_1, \ldots, s_m\}$ is measured. The first version, $\text{ROUGE-L}_{norm}$, is based on the normalized pairwise LCS \citep{radev2002evaluation}, which splits each summary into sentences and matches the sentence pairs with the maximum LCS \citep{lin-2004-rouge}:
\begin{equation}
\label{eq11}
\begin{split}
&\text{ROUGE-L}_{norm} \\
&\;\;= \frac{2 * \sum_{s_i \in S^*} \max_{s_j \in S} \text{LCS}(s_i, s_j)}{|S^*| + |S|} \;,
\end{split}
\end{equation}
where $i$, $j$ refer to sentence indices in $S^*$ and $S$, respectively. $|S*|$ and $|S|$ denote the total number of words in $S^*$ and $S$. To obtain the maximum LCS, sentences are matched regardless of their positions in the summary, and the sentence order within each summary is ignored. As a result, $\text{ROUGE-L}_{norm}$ does not compare the sentence order between two summaries at the entire summary level.

In contrast, the second version, $\text{ROUGE-L}_{full}$, calculates the LCS without splitting the summary into sentences:
\begin{equation}
\label{eq12}
\text{ROUGE-L}_{full} = \frac{2 \, * \, \text{LCS}(S^*, S)}{|S^*| + |S|} \;.
\end{equation}
This version treats each summary as a single complete text, not a set of sentences, and calculates the LCS between the two summaries as a whole \citep{lin-2004-rouge}. This allows for evaluating the order of words or sentences at the entire summary level. Previous studies have commonly used $\text{ROUGE-L}_{norm}$, relying on the \textit{pyrouge} library for evaluation, which only supports $\text{ROUGE-L}_{norm}$. On the other hand, we adopt the \textit{rouge-score} library, developed by Google Research, which implements both versions and allows us to use the second one, $\text{ROUGE-L}_{full}$.

\begin{figure*}[t!]
  \centering
  \includegraphics[width=\linewidth]{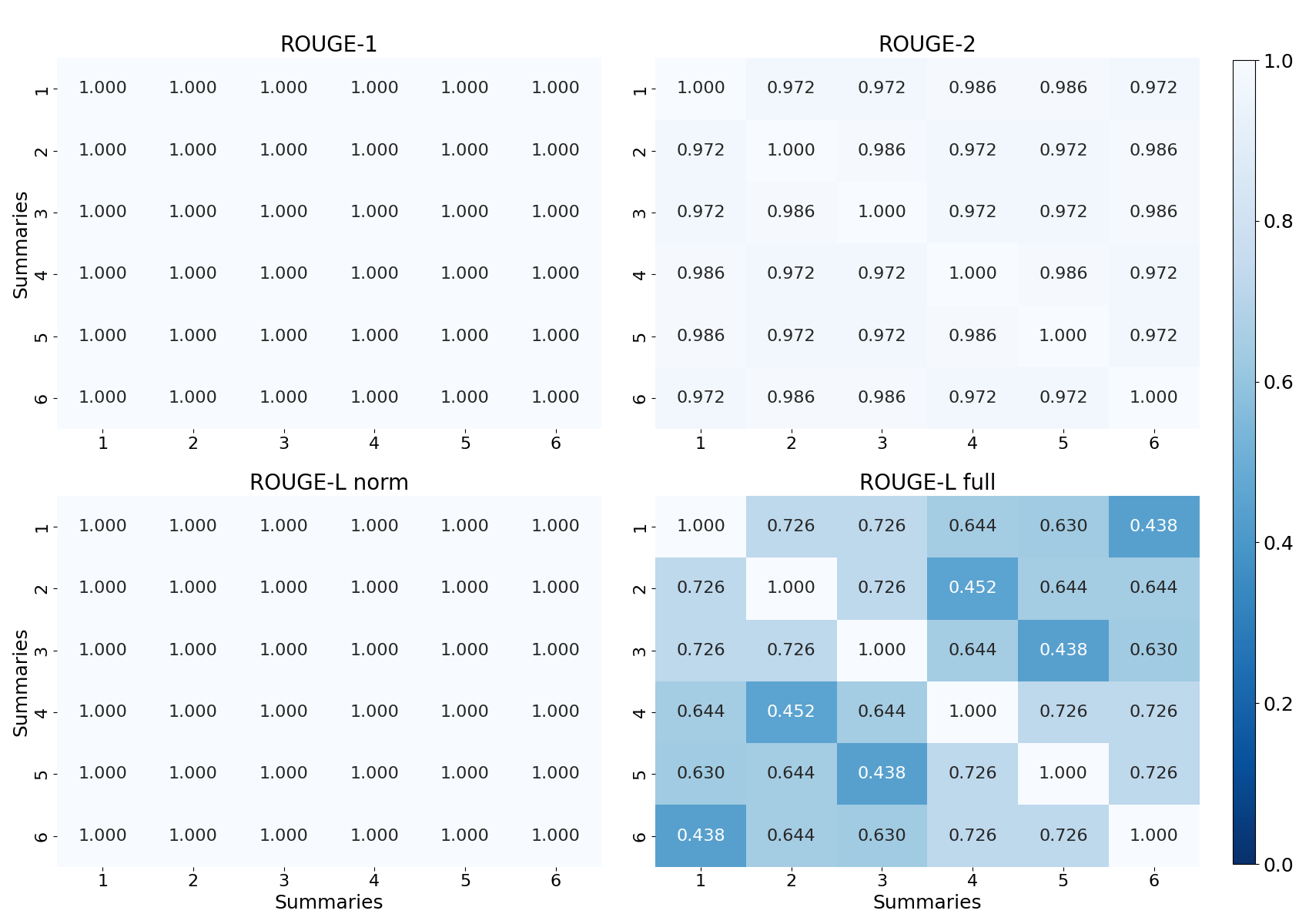}
  \caption{ROUGE-1, ROUGE-2, $\text{ROUGE-L}_{norm}$, and $\text{ROUGE-L}_{full}$ scores for every pair of summaries that share the same sentences but have different sentence orders. The indices of the x-axis and y-axis in each plot indicate each of the six summaries below. If the score between two summaries is close to 1, the ROUGE score hardly detects the difference in sentence order and is not suitable for evaluating sentence order.}
  \label{fig:fig4}
\end{figure*}

To clarify which ROUGE metric is suitable for evaluating sentence order of extractive summaries, we calculate ROUGE-1, ROUGE-2, $\text{ROUGE-L}_{norm}$, and $\text{ROUGE-L}_{full}$ scores for every pair of the six summaries that share the same sentences but differ in their sentence order.
\small
\begin{enumerate}
    \item[0.] [0] The hole was at a point in the plane where weather radars are housed, but the plane landed safely in Denver and no one was injured. [1] A plane was struck by lightning shortly after takeoff during a flight from Reykjavik, Iceland, to Denver, Colorado on Tuesday. [2] It wasn’t until after landing that the passengers and crew found out the lightning strike caused a gaping hole at the nose of the plane.

    \item[1.] [0] The hole was at a point in the plane where weather radars are housed, but the plane landed safely in Denver and no one was injured. [2] It wasn’t until after landing that the passengers and crew found out the lightning strike caused a gaping hole at the nose of the plane. [1] A plane was struck by lightning shortly after takeoff during a flight from Reykjavik, Iceland, to Denver, Colorado on Tuesday.

    \item[2.] [1] A plane was struck by lightning shortly after takeoff during a flight from Reykjavik, Iceland, to Denver, Colorado on Tuesday. [0] The hole was at a point in the plane where weather radars are housed, but the plane landed safely in Denver and no one was injured. [2] It wasn’t until after landing that the passengers and crew found out the lightning strike caused a gaping hole at the nose of the plane.

    \item[3.] [1] A plane was struck by lightning shortly after takeoff during a flight from Reykjavik, Iceland, to Denver, Colorado on Tuesday. [2] It wasn’t until after landing that the passengers and crew found out the lightning strike caused a gaping hole at the nose of the plane. [0] The hole was at a point in the plane where weather radars are housed, but the plane landed safely in Denver and no one was injured.

    \item[4.] [2] It wasn’t until after landing that the passengers and crew found out the lightning strike caused a gaping hole at the nose of the plane. [0] The hole was at a point in the plane where weather radars are housed, but the plane landed safely in Denver and no one was injured. [1] A plane was struck by lightning shortly after takeoff during a flight from Reykjavik, Iceland, to Denver, Colorado on Tuesday.

    \item[5.] [2] It wasn’t until after landing that the passengers and crew found out the lightning strike caused a gaping hole at the nose of the plane. [1] A plane was struck by lightning shortly after takeoff during a flight from Reykjavik, Iceland, to Denver, Colorado on Tuesday. [0] The hole was at a point in the plane where weather radars are housed, but the plane landed safely in Denver and no one was injured.
\end{enumerate}
\normalsize

Figure 4 shows how the four ROUGE metrics detect differences in sentence order. In the first and second plots, ROUGE-1 assigns a score of 1.0 to every pair of summaries, and ROUGE-2 assigns scores ranging from 0.9722 to 0.9861. In the third plot, $\text{ROUGE-L}_{norm}$, another metric for evaluating summarization using ROUGE-L, assigns a score of 1.0 to every pair of summaries. These three metrics hardly detect differences in sentence order, indicating that they are unsuitable for evaluating the sentence order of extractive summaries. In contrast, $\text{ROUGE-L}_{full}$ clearly detects the differences in sentence order between two extractive summaries. Consequently, we adopt $\text{ROUGE-L}_{full}$ as the metric for evaluating sentence order in our experiments.


\section*{Appendix 2}

OrderSum highly increases the number of candidate summaries due to using permutation instead of combination to generate the set of candidate summaries. Because Eq. 8, the objective function for summary reranking in Section 3.5, is computed for every pair of candidate summaries \citep{zhong-etal-2020-extractive, an-etal-2022-colo}, the training cost grows as $O(n^2)$ as the number of candidates increases. To accelerate training with minimal performance degradation, we propose a candidate sampling method called anchor candidate sampling.

The main idea of anchor candidate sampling is that, during training, every case of sentence inclusion is included into a training step for a single document, while the sentence order is learned with varying distributions across the entire training dataset. The equation of anchor candidate sampling is:
\begin{equation}
\label{eq13}
\mathcal{C}_{train} =  \mathcal{C}_A + \text{random} (\mathcal{C}_{NA}, \, |\mathcal{C}_A| (N - 1)).
\end{equation}
$\mathcal{C}_{train}$ is the subset of entire candidate summaries $\mathcal{C}$ sampled for training. $ \mathcal{C}_A$ denotes anchor candidates, which are candidate summaries generated by combination. $\mathcal{C}_{NA}$ denotes non-anchor candidates, which are the remaining subset of candidate summaries excluding $\mathcal{C}_A$. $\mathcal{C}_{NA}$ contains candidate summaries that share the same sentences as $\mathcal{C}_A$ but have different sentence orders. $N$ is the sampling factor that determines the number of $\mathcal{C}_{NA}$ involved in $\mathcal{C}_{train}$. $N$ is set as a multiplier for the size of $\mathcal{C}_A$ considering the ratio between $\mathcal{C}_A$ and $\mathcal{C}_{NA}$.

\begin{table}[t!]
  \centering
  \setlength{\tabcolsep}{11pt} 
  \renewcommand{\arraystretch}{1.35} 
  \small
    \begin{tabular}{lccc}
      \toprule
      \textbf{} & \textbf{R-1} & \textbf{R-2} & \textbf{R-L}\\
      
      \midrule
      \multicolumn{4}{l}{\textbf{WikiHow $\boldsymbol{(300,\; 16)}$}} \\ 
      \hline
      \textbf{$\boldsymbol{N=2 \; (32)}$\;} & \textbf{32.98} & \textbf{9.69} & 21.05 \\
      \textbf{$\boldsymbol{N=4 \; (64)}$\;} & 32.86 & 9.63 & \textbf{21.11} \\
      \textbf{$\boldsymbol{N=8 \; (128)}$\;} & 31.64 & 9.22 & 20.36 \\
    
      \midrule
      \multicolumn{4}{l}{\textbf{PubMed $\boldsymbol{(60,480,\; 36)}$}} \\ 
      \hline
      \textbf{$\boldsymbol{N=2 \; (72)}$\;} & \textbf{41.58} & \textbf{14.64} & \textbf{21.19} \\
      \textbf{$\boldsymbol{N=4 \; (144)}$\;} & 41.33 & 14.54 & 21.01 \\
      \textbf{$\boldsymbol{N=8 \; (288)}$\;} & 40.20 & 13.95 & 20.48 \\
      
      \bottomrule 
    \end{tabular} 
  \normalsize
  \caption{Ablation studies on different sampling factors of anchor candidate sampling. We conduct experiments on two datasets with a large number of candidate summaries, WikiHow and PubMed. We show the number of entire candidate summaries $\mathcal{C}$ and anchor candidates $\mathcal{C}_A$ for each dataset. We show the number of candidate summaries used for training $\mathcal{C}_{train}$ corresponding to each sampling factor.}
  \label{table:9}
\end{table}

Table 9 is the ablation studies for anchor candidate sampling with different sampling factors. The results show that anchor candidate sampling enables the model to learn sentence order effectively with a small sampling factor. On WikiHow, the model trained with $N = 2$ obtains the highest ROUGE-1 and ROUGE-2 scores, and the model trained with $N = 4$ obtains the highest ROUGE-L score, using only 32 and 64 among 300 candidate summaries, respectively. On PubMed, the model trained with $N = 2$ achieves the highest scores for ROUGE-1, ROUGE-2, and ROUGE-L, using only 72 among 60,480 candidate summaries. In contrast, the models trained with $N = 8$ obtain the lowest scores across all ROUGE scores on both datasets. These results demonstrate that anchor candidate sampling dramatically reduces the number of candidate summaries and training costs while maintaining performance in both sentence inclusion and sentence order.


\section*{Appendix 3}

Figure 5 shows the validation graphs obtained during the training of BARTSUM, CoLo, and OrderSum on the four datasets. As explained in Section 5.4, OrderSum continuously increases the ROUGE-L scores during training on the four datasets. This indicates that OrderSum trains sentence order using a new objective function that includes ROUGE-L, unlike the previous models.

For the remaining metrics, ROUGE-1 and ROUGE-2, the validation graphs for CoLo and OrderSum show similar patterns, saturating early from the starting point initialized with BARTSUM.

\begin{figure*}[t!]
  \centering
  \includegraphics[width=\linewidth]{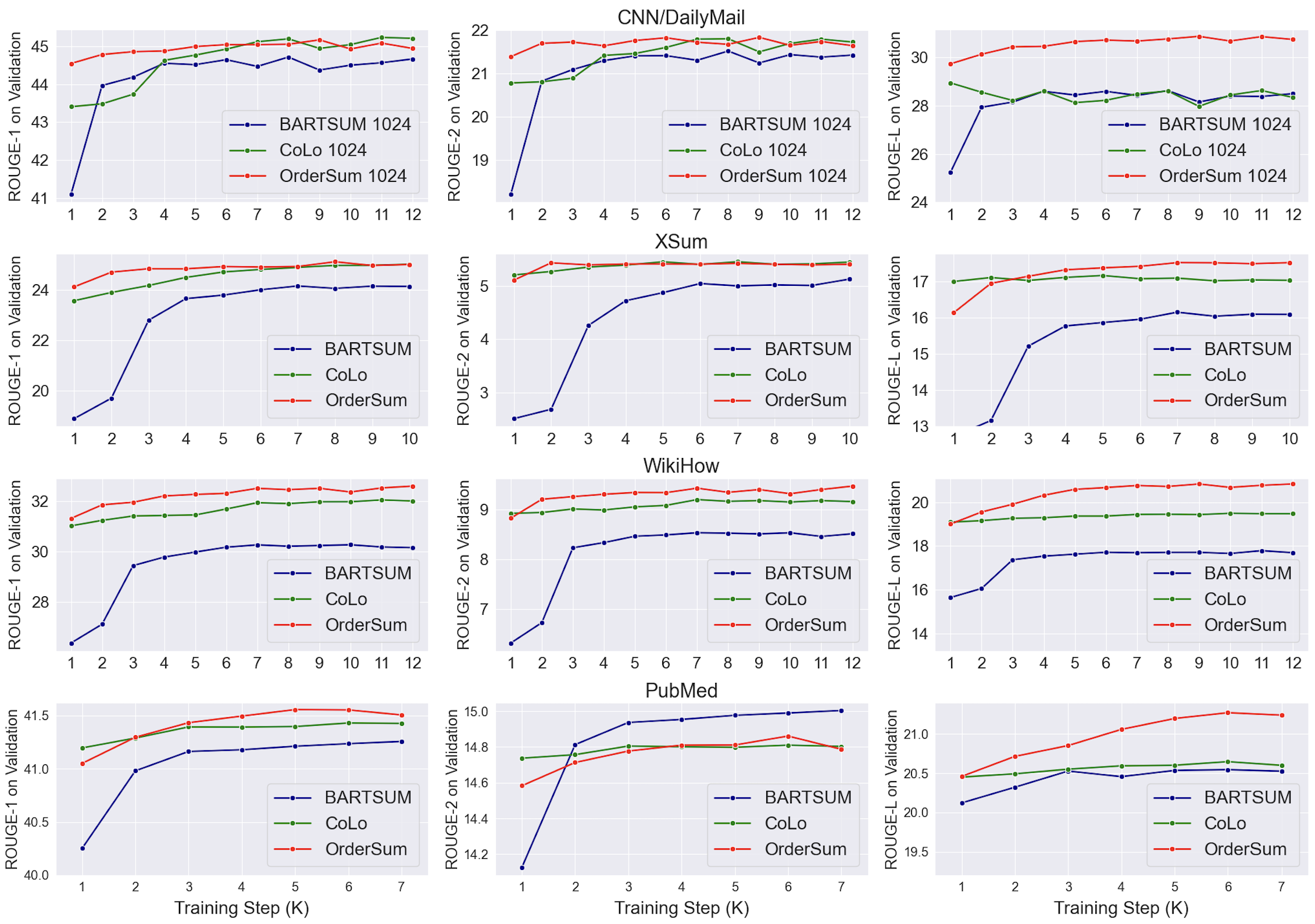}
  \caption{Validation graphs for ROUGE-1, ROUGE-2, and ROUGE-L scores during the training of BARTSUM, CoLo, and OrderSum on CNN/DailyMail, XSum, WikiHow, and PubMed. On CNN/DailyMail, BARTSUM 1024, CoLo 1024, and OrderSum 1024 are used to obtain the graphs. BARTSUM, CoLo, and OrderSum are used on the remaining three datasets. Training is conducted for 12K, 10K, 12K, and 7K steps on CNN/DailyMail, XSum, WikiHow, and PubMed, respectively, with validation performed every 1,000 steps.}
  \label{fig:fig5}
\end{figure*}


\section*{Appendix 4}

Table 10 illustrates the system prompt and user prompt given to GPT-4o for evaluating LLM performance on the extractive summarization task. We are motivated by \citet{zhang-etal-2023-extractive-summarization} when designing the instruction on extractive summarization for the LLM. To enhance the capability of the LLM, we provide an instruction in the system prompt consisting of two kinds of information: 1) an explanation of the extractive summarization task and 2) three-shot examples, each containing a document, a reference summary, and an extractive summary corresponding to Ordered ORACLE. The examples are obtained from the training set of each dataset. We prepend indices starting from 0 before each sentence within a document in both the system and user prompts to minimize hallucination in predictions. The extractive summaries in the examples avoid cases where the sentence indices of the prediction are overly concentrated at the beginning of the document or where the sentence indices are sorted in ascending order.


\section*{Appendix 5}

Tables 11, 12, 13, and 14 show example summaries generated by CoLo \citep{an-etal-2022-colo} and OrderSum on the CNN/DailyMail \citep{NIPS2015_afdec700}, XSum \citep{narayan-etal-2018-dont}, WikiHow \citep{koupaee2018wikihow}, and PubMed \citep{cohan-etal-2018-discourse} test sets, respectively. The examples show that OrderSum improves sentence order and the quality of summaries not only on CNN/DailyMail but also on the other three datasets, compared to CoLo. In particular, on PubMed, OrderSum significantly changes the sentence order and achieves better alignment with the reference summaries compared to CoLo, unlike the results in Table 4, where OrderSum increases ROUGE-L by a relatively small margin.

The numbers at the top of each example indicate the index within the test set. In Table 11, the example summaries are generated by CoLo 1024 and OrderSum 1024.

\onecolumn

\begin{table*}[t!]
  \centering
  \setlength{\tabcolsep}{12pt} 
  \renewcommand{\arraystretch}{1.3} 
  \small
  \begin{tabular}{p{0.945\linewidth}}
    \toprule
    \textbf{System Prompt} \\
    \hline
    You are an expert in extractive summarization tasks. Your role is to extract 2-3 sentences from a given document to create accurate and complete extractive summaries. \\
    
    Examples of good extractive summaries are given below. In the examples, the text between <document> tags is the original document from which sentences are extracted. Each sentence in the document is preceded by a sentence index following the existing extractive summarization datasets. The text between <reference\_summary> tags is the human-written summary of the given document. The reference summary serves as the target of the extractive summary. In the examples, the list of sentence indices between <extractive\_summary> tags is the correct prediction for extractive summarization on the given document. The extractive summary created by arranging the sentences of each index represents the closest match to the reference summary from the given document. \\
    <example\_1> \\
    <document> \\
    $\text{[0]}$ Didier Drogba is poised to take a step closer to realising his dream .... $\text{[1]}$ Talks have been ongoing over a one year player-coach contract .... $\text{[2]}$ Drogba is understood to have already texted some of his former Chelsea teammates .... \\
    </document> \\
    <reference\_summary> \\
    The 36-year-old Chelsea legend has interest from Gatar and Juventus. Mourinho has a strong relationship with Drogba .... \\
    </reference\_summary> \\
    <extractive\_summary> \\
    (5, 2, 22) \\
    </extractive\_summary> \\
    </example\_1> \\

    <example\_2> \\
    .... \\
    </example\_2> \\

    <example\_3> \\
    .... \\
    </example\_3> \\
    Understand the patterns of good extractive summaries from the given examples. When the user gives a new document, predict the list of sentence indices for the extractive summary. You don't necessarily need to predict the sentences in ascending order. Determine the sentence order to make the extractive summary flow naturally, so that it matches the human-written summary as closely as possible. \\
    
    \midrule
    \textbf{User Prompt} \\
    \hline
    <document> \\
    $\text{[0]}$ Passengers of an Icelandair flight to Denver on Tuesday were left in a bit of shock after their plane was struck by lightning shortly after takeoff. $\text{[1]}$ Though the Boeing 757 continued its 3,700-mile journey from Reykjavik, Iceland, to Denver, .... \\
    </document> \\
    <extractive\_summary> \\
    
    \bottomrule 
    \end{tabular} 
  \normalsize
  \caption{Example of the system prompt and the user prompt given to the "gpt-4o-2024-08-06" version of GPT-4o [39] performing extractive summarization on CNN/DailyMail.}
  \label{table:10}
\end{table*}

\clearpage

\begin{longtable}{l>{\raggedright\arraybackslash}p{0.8\textwidth}}
    \captionsetup{font=small}
    \caption{Examples from the CNN/DailyMail test set.} \label{table:11} \\
    
    \toprule
    \multicolumn{2}{c}{\small \textbf{CNN/DailyMail}} \\
    \midrule
    \endfirsthead
    
    \toprule
    \multicolumn{2}{c}{\small \textbf{CNN/DailyMail}} \\
    \midrule
    \endhead

    \endfoot
    \bottomrule
    \endlastfoot

    \small \textbf{\# 255} & \\
    \cmidrule(lr){1-2}
    \small \textbf{Document} & \small greece has demanded more than 200 billion in compensation from germany for nazi atrocities during the second world war . the government yesterday unveiled its final calculation for the war reparations stemming from occupation by the third reich . the radical left syriza party says germany owes greece nearly 279 billion euros , or 204 billion to compensate it for looting and war crimes . greek prime minister alexis tsipras raised the reparations issue when he met german chancellor angela merkel in berlin last month .... \\
    
    \cmidrule(lr){1-2}
    \small \textbf{Reference} & \small greek government has unveiled its final calculation for the war reparations . radical left syriza party says germany owes greece nearly 279 billion euros . the german government claims the issue was resolved legally years ago . it comes days before greece is obliged to pay off 450million euros of debt . \\
    
    \cmidrule(lr){1-2}
    \small \textbf{CoLo 1024} & \small greece suffered a brutal occupation at the hands adolf hitler 's forces in 1941 . the german government says the issue was resolved legally years ago . the radical left syriza party says germany owes greece nearly 279 billion euros , or 204 billion to compensate it for looting and war crimes . \\
    
    \cmidrule(lr){1-2}
    \small \textbf{OrderSum 1024} & \small the radical left syriza party says germany owes greece nearly 279 billion euros , or 204 billion to compensate it for looting and war crimes . the german government says the issue was resolved legally years ago . the demand comes just days before greece is obliged to pay 450 million euros of its debt to the international monetary fund . \\

    \midrule
    \small \textbf{\# 1532} & \\
    \cmidrule(lr){1-2}
    \small \textbf{Document} & \small the queensland government has called a government decision to deport a single-mother and her son cold and heartless ' . townsville nurse maria sevilla and her son , tyrone , will be forced out of the country within 28 days , after ms sevilla 's appeal to the immigration department was unsuccessful . the single-mum initially had her visa application rejected because her son 's autism would be a burden on the australian health system ' .... \\
    
    \cmidrule(lr){1-2}
    \small \textbf{Reference} & \small mother and autistic son to be deported to philippines within 28 days . maria sevilla and her son , tyrone , have lived in australia since 2007 . appeal rejected because son 's autism could be a burden on government . their family is in queensland , sevilla is a nurse at townsville hospital . queensland disabilities minister says decision is ` cold and heartless ' . immigration department says child does not meet visa health requirements . \\
    
    \cmidrule(lr){1-2}
    \small \textbf{CoLo 1024} & \small the single-mum initially had her visa application rejected because her son 's autism would be a ` burden on the australian health system ' , according to the immigration department . after ms sevilla 's appeal was unsuccessful , her and her son will be deported within 28 days . ms sevilla , a single mum and nurse in the queensland town of townsville , has lived in australia with her son since 2007 . \\
    
    \cmidrule(lr){1-2}
    \small \textbf{OrderSum 1024} & \small townsville nurse maria sevilla and her son , tyrone , will be forced out of the country within 28 days , after ms sevilla 's appeal to the immigration department was unsuccessful . the single-mum initially had her visa application rejected because her son 's autism would be a burden on the australian health system ' , according to the immigration department . the queensland government has called a government decision to deport a single-mother and her son cold and heartless ' . \\

    \midrule
    \small \textbf{\# 10317} & \\
    \cmidrule(lr){1-2}
    \small \textbf{Document} & \small a fashion designer 's 30,000 sq ft los angeles mansion , which features a two-story-tall chandelier and 22 bathrooms , is now on the market with a whopping \$ 85 million price tag . max azria 's home is spread across three acres in holmby hills , one of the three affluent neighborhoods that makes up the city 's platinum triangle ' and bordered by beverly hills and bel air . visitors to the sprawling home , dubbed maison du solei .... \\
    
    \cmidrule(lr){1-2}
    \small \textbf{Reference} & \small max azria 's 30,000 sq ft home is spread across three acres and boasts 60 rooms . there is also a glass-enclosed tennis court with its own viewing box and five different gardens . the zero-edged swimming pool has a moroccan-style bathhouse , complete with a sauna and spa . azria and his wife lubov bought the house for \$ 14.4 m in 2005 before giving it a \$ 30 m renovation . \\
    
    \cmidrule(lr){1-2}
    \small \textbf{CoLo 1024} & \small there are 17 bedrooms to choose from and 60 rooms to explore , and that 's not even counting the guesthouse and 6,000 sq ft movie theater , complete with its own popcorn and snack bar . visitors to the sprawling home , dubbed maison du solei , are first greeted by a floor-to-ceiling waterfall chandelier made up of 150,000 crystals . the home is spread across three acres in holmby hills , one of the three affluent neighborhoods that makes up the city 's ` platinum triangle .' \\
    
    \cmidrule(lr){1-2}
    \small \textbf{OrderSum 1024} & \small max azria 's home is spread across three acres in holmby hills , one of the three affluent neighborhoods that makes up the city 's ` platinum triangle ' and bordered by beverly hills and bel air . visitors to the sprawling home , dubbed maison du solei , are first greeted by a floor-to-ceiling waterfall chandelier made up of 150,000 crystals . there are 17 bedrooms to choose from and 60 rooms to explore , and that 's not even counting the guesthouse and 6,000 sq ft movie theater , complete with its own popcorn and snack bar . \\

\end{longtable}


\clearpage
\begin{longtable}{l>{\raggedright\arraybackslash}p{0.8\textwidth}}
    \captionsetup{font=small}
    \caption{Examples from the XSum test set.} \label{table:12} \\
    
    \toprule
    \multicolumn{2}{c}{\small \textbf{XSum}} \\
    \midrule
    \endfirsthead
    
    \toprule
    \multicolumn{2}{c}{\small \textbf{XSum}} \\
    \midrule
    \endhead

    \endfoot
    \bottomrule
    \endlastfoot

    \small \textbf{\# 265} & \\
    \cmidrule(lr){1-2}
    \small \textbf{Document} & \small the body of lindsay rimer , 13 , was found in the rochdale canal five months after she disappeared in hebden bridge , west yorkshire , in november 1994 . her killer has never been found and west yorkshire police are treating the death as murder . the force said a new dna profile had been identified , which it hoped would lead them to identify the killer . speaking on the anniversary of the discovery of her body , det supt simon atkinson said .... \\
    
    \cmidrule(lr){1-2}
    \small \textbf{Reference} & \small new forensic leads are being investigated in connection with the death of a schoolgirl 21 years ago . \\
    
    \cmidrule(lr){1-2}
    \small \textbf{CoLo} & \small he added : '' it 's a leap forward we have n't had in the past 21 years , so i 'm really excited about this development . '' the force said a new dna profile had been identified , which it hoped would lead them to identify the killer . \\
    
    \cmidrule(lr){1-2}
    \small \textbf{OrderSum} & \small the force said a new dna profile had been identified , which it hoped would lead them to identify the killer . her killer has never been found and west yorkshire police are treating the death as murder . \\

    \midrule
    \small \textbf{\# 5248} & \\
    \cmidrule(lr){1-2}
    \small \textbf{Document} & \small he players , who thrashed japan 5-2 on sunday in vancouver , will ride an open-topped bus down a stretch of broadway commonly known as the canyon of heroes . it 's a rare honour for athletes that do not come from new york , but pressure has been building in recent days to honour their achievement . the game was the most-watched football match in us history . a 13-minute hat-trick by carli lloyd blew japan away early in the match , watched by an average of 25.4 million viewers .... \\
    
    \cmidrule(lr){1-2}
    \small \textbf{Reference} & \small a ticker-tape parade in lower manhattan on friday will honour the usa 's world cup-winning women 's football team . \\
    
    \cmidrule(lr){1-2}
    \small \textbf{CoLo} & \small president barack obama was among those to pay tribute to their success , which was the third time the usa women 's team had won the world cup . a rally was held in los angeles for the team on tuesday , and hours later the parade was announced by new york mayor bill de blasio . \\
    
    \cmidrule(lr){1-2}
    \small \textbf{OrderSum} & \small the last time the city honoured a team of national athletes was in 1984 after the los angeles olympics . president barack obama was among those to pay tribute to their success , which was the third time the usa women 's team had won the world cup . \\

    \midrule
    \small \textbf{\# 8326} & \\
    \cmidrule(lr){1-2}
    \small \textbf{Document} & \small set 302 to win from 46 overs at taunton , middlesex completed a dramatic run chase with two balls remaining . if we were serious about winning the title , we want to win games . you ca n't just keep playing for draws and expect to be up there , '' he told bbc somerset . we are here to provide entertainment . we came up short but gave everything . '' defeat left somerset sixth in the county championship division one table with 96 points , 45 behind league leaders middlesex .... \\
    
    \cmidrule(lr){1-2}
    \small \textbf{Reference} & \small somerset captain chris rogers says he has `` no regrets '' about his declaration against middlesex on wednesday , despite losing the match by two wickets . \\
    
    \cmidrule(lr){1-2}
    \small \textbf{CoLo} & \small `` if we were serious about winning the title , we want to win games . somerset took eight wickets as middlesex chased their target on wednesday , but a six from wicketkeeper john simpson in the final over won the match . \\
    
    \cmidrule(lr){1-2}
    \small \textbf{OrderSum} & \small `` hindsight is a wonderful thing , '' former australia test opener rogers added . somerset took eight wickets as middlesex chased their target on wednesday , but a six from wicketkeeper john simpson in the final over won the match . \\

\end{longtable}


\clearpage
\begin{longtable}{l>{\raggedright\arraybackslash}p{0.8\textwidth}}
    \captionsetup{font=small}
    \caption{Examples from the WikiHow test set.} \label{table:13} \\
    
    \toprule
    \multicolumn{2}{c}{\small \textbf{WikiHow}} \\
    \midrule
    \endfirsthead
    
    \toprule
    \multicolumn{2}{c}{\small \textbf{WikiHow}} \\
    \midrule
    \endhead

    \endfoot
    \bottomrule
    \endlastfoot

    \small \textbf{\# 724} & \\
    \cmidrule(lr){1-2}
    \small \textbf{Document} & \small start by doing simple stretches like neck rolls and shoulder rolls . to do shoulder rolls , sit or stand up straight . then , roll your shoulder up , back , and down in a fluid motion . repeat this movement five to 10 times . then reverse it , rolling forward five to 10 times . this should help to loosen up your shoulder muscles to do neck rolls , sit or stand up straight with your shoulders rolled back . slowly tilt your head to the right and roll it down , chin to chest .... \\
    
    \cmidrule(lr){1-2}
    \small \textbf{Reference} & \small do shoulder rolls and neck rolls . try the thread the needle pose . do cat and cow pose . practice bridge pose . try a standing forward fold . do extended side angle pose . \\
    
    \cmidrule(lr){1-2}
    \small \textbf{CoLo} & \small repeat cat and cow pose for five to 10 breaths total . bridge pose is another great yoga posture for stretching and opening the shoulders . start by doing simple stretches like neck rolls and shoulder rolls . you can do a forward fold on an exercise mat or even on the carpeted floor in your office to do a standing forward fold , stand up straight with your legs hip-width apart . \\
    
    \cmidrule(lr){1-2}
    \small \textbf{OrderSum} & \small start by doing simple stretches like neck rolls and shoulder rolls . repeat cat and cow pose for five to 10 breaths total . bridge pose is another great yoga posture for stretching and opening the shoulders . forward folds are great for releasing tension and stress in the shoulder muscles . \\

    \midrule
    \small \textbf{\# 1349} & \\
    \cmidrule(lr){1-2}
    \small \textbf{Document} & \small if you are someone who felt defined by your career , you may experience a significant sense of loss upon your retirement . also , the thought of somebody replacing you at work can make you feel unimportant . it can be difficult to validate yourself without the connection to a career . over time , you may begin to feel depressed or anxious , and you may even begin to second-guess your decision to retire. build a new identity for yourself by using your time to find meaningful ways to connect with other people .... \\
    
    \cmidrule(lr){1-2}
    \small \textbf{Reference} & \small anticipate the emotional impact of giving up your career . adjust to spending more time with your spouse or family . decide how you will structure your days . find other sources of social interaction . \\
    
    \cmidrule(lr){1-2}
    \small \textbf{CoLo} & \small whichever you are , you will need to schedule activities that give you social interactions . it can be difficult to validate yourself without the connection to a career . spend more time being active . \\
    
    \cmidrule(lr){1-2}
    \small \textbf{OrderSum} & \small it can be difficult to validate yourself without the connection to a career . remember that with time you will learn to adjust to spending more time at home with your spouse . whichever you are , you will need to schedule activities that give you social interactions . \\

    \midrule
    \small \textbf{\# 5672} & \\
    \cmidrule(lr){1-2}
    \small \textbf{Document} & \small you can use any type of fabric you want , but cotton , linen , or canvas would be best , because they are durable.you can make this bag bigger / smaller . use the same measurements for the casing and the seam allowances , however, place one of the pieces of fabric down , with the wrong side facing you . fold both of the long edges and one of the narrow edges down by 1/4 inch ( 0.64 centimeter ) . press them flat with an iron ; be sure to use a heat setting suitable for the fabric you are working with .... \\
    
    \cmidrule(lr){1-2}
    \small \textbf{Reference} & \small cut two 12 by 14-inch ( 30.48 by 35.56-centimeter ) rectangles out of fabric . fold and press three of the edges down by 1/4 inch ( 0.64 centimeter ) . top stitch the three folded edges down using a sewing machine . make the casing by folding the top , raw edge down twice . topstitch the casing down , as close to the bottom edge as possible . sew the two fabric pieces together . turn the bag inside out , and press it flat with an iron . thread the cords through the casings . knot the cords together . turn the bag inside out and pull the cords through the gaps you left earlier at the bottom corners . stitch the gaps close . turn the bag right-side-out . \\
    
    \cmidrule(lr){1-2}
    \small \textbf{CoLo} & \small take the two cords on the left side of the bag , and knot the ends together . repeat this with the two cords on the right side of the bag. , pull the left cord through the left gap , and the right cord through the right gap . to close the back , simply scrunch the top of the bag shut along the cord . fold both of the long edges and one of the narrow edges down by 1/4 inch ( 0.64 centimeter ) . starting just below the casing , sew all the way around the side , bottom , and side edges of the bag , using a 5/8-inch ( 0.59-centimeter ) seam allowance.leave a 1-inch ( 2.54-centimeter ) wide gap at each bottom corner . \\
    
    \cmidrule(lr){1-2}
    \small \textbf{OrderSum} & \small fold both of the long edges and one of the narrow edges down by 1/4 inch ( 0.64 centimeter ) . do not sew the two fabric pieces together just yet. , fold the top , raw edge down by 1/2 inch ( 1.27 centimeters ) and press it flat with an iron . repeat this with the two cords on the right side of the bag. , pull the left cord through the left gap , and the right cord through the right gap . starting just below the casing , sew all the way around the side , bottom , and side edges of the bag , using a 5/8-inch ( 0.59-centimeter ) seam allowance.leave a 1-inch ( 2.54-centimeter ) wide gap at each bottom corner . take the two cords on the left side of the bag , and knot the ends together . \\

\end{longtable}


\clearpage
\begin{longtable}{l>{\raggedright\arraybackslash}p{0.8\textwidth}}
    \captionsetup{font=small}
    \caption{Examples from the PubMed test set.} \label{table:14} \\
    
    \toprule
    \multicolumn{2}{c}{\small \textbf{PubMed}} \\
    \midrule
    \endfirsthead
    
    \toprule
    \multicolumn{2}{c}{\small \textbf{PubMed}} \\
    \midrule
    \endhead

    \endfoot
    \bottomrule
    \endlastfoot

    \small \textbf{\# 168} & \\
    \cmidrule(lr){1-2}
    \small \textbf{Document} & \small uterine adenomyosis is a common gynecologic disorder characterized by the presence and growth of heterotopic endometrial or endometrium - like structures in the myometrium , with adjacent smooth muscle hyperplasia and can lead to dysmenorrhea and infertility . the ectopic endometrial tissue induces hypertrophy and hyperplasia of the surrounding myometrium , resulting in diffuse globular enlargement of the uterus analogous to the concentric enlargement of the pregnant uterus . the presenting symptoms include a soft and diffusely enlarged uterus with menorrhagia ( 40 \% 50 \% ) , dysmenorrhea ( 10 \% 30 \% ) , metrorrhagia ( 10 \% 12 \% ), dyspareunia and dyschezia . the advised treatment for severe adenomyosis is total hysterectomy , .... \\
    
    \cmidrule(lr){1-2}
    \small \textbf{Reference} & \small adenomyosis is a common gynecological disorder characterized by the presence of endometrial glands and stroma deep within the myometrium associated with myometrial hypertrophy and hyperplasia . focal uterine infarction after ivf - et in a patient with adenomyosis following biochemical pregnancy has not been previously reported , although it occurs after uterine artery embolization in order to control symptoms caused by fibroids or adenomyosis . we report a case of a nulliparous woman who had uterine adenomyosis presenting with fever , pelvic pain and biochemical abortion after undergoing an ivf - et procedure and the detection of a slightly elevated serum hcg . focal uterine infarction was suspected after a pelvic magnetic resonance imaging demonstrated preserved myometrium between the endometrial cavity and inner margin of the necrotic myometrium . this case demonstrates that focal uterine infarction should be considered in the differential diagnosis of acute abdominal pain , vaginal bleeding and infectious signs in women experiencing biochemical abortion after an ivf - et procedure . \\
    
    \cmidrule(lr){1-2}
    \small \textbf{CoLo} & \small this is the first report of uterine infarction after ivf - et in a patient with uterine adenomyosis in korea . although uterine infarction is a relatively common occurrence after uae for the treatment of fibroids or adenomyosis , uterine infarction after ivf - et in a patient with adenomyosis is very rare . to our knowledge . the advised treatment for severe adenomyosis is total hysterectomy , but for patients wishing to preserve their uterus , a minimally invasive alternative procedure , uterine artery embolization ( uae ) , can be performed . uae is a nonsurgical alternative for patients with menorrhagia , symptomatic adenomyosis , or symptomatic uterine fibroids . uterine adenomyosis is a common gynecologic disorder characterized by the presence and growth of heterotopic endometrial or endometrium - like structures in the myometrium , with adjacent smooth muscle hyperplasia and can lead to dysmenorrhea and infertility . the presenting symptoms include a soft and diffusely enlarged uterus with menorrhagia ( 40 \% 50 \% ) , dysmenorrhea ( 10 \% 30 \% ) , metrorrhagia ( 10 \% 12 \% ) , dyspareunia and dyschezia . \\
    
    \cmidrule(lr){1-2}
    \small \textbf{OrderSum} & \small uterine adenomyosis is a common gynecologic disorder characterized by the presence and growth of heterotopic endometrial or endometrium - like structures in the myometrium , with adjacent smooth muscle hyperplasia and can lead to dysmenorrhea and infertility . the ectopic endometrial tissue induces hypertrophy and hyperplasia of the surrounding myometrium , resulting in diffuse globular enlargement of the uterus analogous to the concentric enlargement of the pregnant uterus . , this is the first report of uterine infarction after ivf - et in a patient with uterine adenomyosis in korea . the advised treatment for severe adenomyosis is total hysterectomy , but for patients wishing to preserve their uterus , a minimally invasive alternative procedure , uterine artery embolization ( uae ) , can be performed . although uterine infarction is a relatively common occurrence after uae for the treatment of fibroids or adenomyosis , uterine infarction after ivf - et in a patient with adenomyosis is very rare . to our knowledge . uae is a nonsurgical alternative for patients with menorrhagia , symptomatic adenomyosis , or symptomatic uterine fibroids . \\

    \midrule
    \\
    \\
    \small \textbf{\# 1542} & \\
    \cmidrule(lr){1-2}
    \small \textbf{Document} & \small osteoporosis most commonly affects postmenopausal women , placing them at a significant risk of fractures . in particular , hip fractures are an important cause of mortality and morbidity among postmenopausal women . approximately 20 \% of patients with hip fractures die within a year , most of the deaths occurring within the first 6 months after a fracture ( cumming et al 1997 ) . among the survivors , recent reports have shown that the rate of hip fractures is falling in ontario and finland ( jaglal et al 2005 ; kannus et al 2006 ) .... \\
    
    \cmidrule(lr){1-2}
    \small \textbf{Reference} & \small osteoporosis most commonly affects postmenopausal women , placing them at a significant risk of fractures . in particular , hip fractures are an important cause of mortality and morbidity among postmenopausal women . anti - resorptive therapies that produce greater decreases in bone turnover markers together with greater increases in bone mineral density ( bmd ) are associated with greater reductions in fracture risk , especially at sites primarily composed of cortical bone such as the hip . thus , treatment with potent anti - resorptive drugs like alendronate is a strategy for preventing hip fractures in postmenopausal women with osteoporosis . the purpose of this paper is to discuss the efficacy of alendronate against hip fractures and the mechanism for this anti - fracture efficacy in postmenopausal women with osteoporosis . a meta - analysis of randomized controlled trials has shown that alendronate reduces the risk of hip fractures by 55 \% in postmenopausal women with osteoporosis . according to the analyses of the fracture intervention trial , each 1 standard deviation reduction in a 1-year change in bone - specific alkaline phosphatase ( bsap ) is associated with 39 \% fewer hip fractures in alendronate - treated postmenopausal women , and those with at least 30 \% reduction in bsap have a 74 \% lower risk of hip fractures relative to those with less than 30 \% . alendronate is effective in reducing the risk of hip fractures across a spectrum of ages . the mechanism for this anti - fracture efficacy has been clarified ; alendronate strongly suppresses bone turnover and subsequently increases hip bmd , decreases cortical porosity , improves parameters of hip structure geometry ( cortical thickness , cross - sectional area , section modulus , and buckling ratio ) , and produces more uniform mineralization ( increases the mean degree of mineralization of bone ) in cortical bone . a once - weekly regimen of alendronate administration provides better patient compliance and persistence with the treatment than the once - daily dosing regimen , leading to greater efficacy against hip fractures . thus , the efficacy of alendronate against hip fractures has been confirmed in postmenopausal women with osteoporosis , especially with a once - weekly dosing regimen . \\
    
    \cmidrule(lr){1-2}
    \small \textbf{CoLo} & \small thus , this paper discusses , based on a review of the literature , the efficacy of alendronate against hip fractures and the mechanism for this anti - fracture efficacy in postmenopausal women with osteoporosis . treatment with potent anti - resorptive drugs is a strategy for preventing hip fractures in postmenopausal women . osteoporosis most commonly affects postmenopausal women , placing them at a significant risk of fractures . in particular , hip fractures are an important cause of mortality and morbidity among postmenopausal women . in particular , evidence to explain the efficacy of alendronate against hip fractures has been accumulated for postmenopausal osteoporosis . alendronate and risedronate , which are the second- and third - generation bisphosphonates , respectively , have been used as first - line drugs in the treatment of osteoporosis for many years . increase in hip bmd may be necessary to decrease the fracture risk of the hip , which is primarily composed of cortical bone and may require greater proportionate changes than trabecular bone ( epstein 2007 ) . \\
    
    \cmidrule(lr){1-2}
    \small \textbf{OrderSum} & \small osteoporosis most commonly affects postmenopausal women , placing them at a significant risk of fractures . in particular , hip fractures are an important cause of mortality and morbidity among postmenopausal women . thus , this paper discusses , based on a review of the literature , the efficacy of alendronate against hip fractures and the mechanism for this anti - fracture efficacy in postmenopausal women with osteoporosis . in particular , evidence to explain the efficacy of alendronate against hip fractures has been accumulated for postmenopausal osteoporosis . treatment with potent anti - resorptive drugs is a strategy for preventing hip fractures in postmenopausal women . is multiplied when both are present ( garnero et al 1996 ) , bone turnover and bmd are important factors in decreasing the risk of hip fractures . both reduction in bone turnover and increase in hip bmd may be necessary to decrease the fracture risk of the hip , which is primarily composed of cortical bone and may require greater proportionate changes than trabecular bone ( epstein 2007 ) . alendronate and risedronate , which are the second- and third - generation bisphosphonates , respectively , have been used as first - line drugs in the treatment of osteoporosis for many years . \\

    \midrule
    \small \textbf{\# 4296} & \\
    \cmidrule(lr){1-2}
    \small \textbf{Document} & \small catel manzke syndrome ( mim : 616145 ) is a rare genetic disorder characterized by pierre robin sequence with hyperphalangy and clinodactyly of the index finger , . affected individuals typically present with the classic features of pierre robin sequence including micrognathia , airway obstruction secondary to displacement of the tongue base , and often , cleft palate . the orthopedic abnormalities are the result of an accessory bone between the second metacarpal and proximal phalanx that causes radial deviation of the second digit ( manzke dysostosis ) .... \\
    
    \cmidrule(lr){1-2}
    \small \textbf{Reference} & \small catel manzke syndrome is a rare autosomal recessive disorder characterized by pierre robin sequence with hyperphalangy and clinodactyly of the index finger . recently , homozygous or compound heterozygous pathogenic variants in tgds have been discovered to cause catel manzke syndrome . here , we describe a 12-month - old male with molecularly confirmed catel manzke syndrome who presented with pierre robin sequence ( but without cleft palate ) and hyperphalangy , and we compare his phenotype with the seven previously described patients with pathogenic variants in tgds . our patient is on the severe end of the phenotypic spectrum , presenting with respiratory complications and failure to thrive . furthermore , our finding of a homozygous p.ala100ser pathogenic variant in our patient supports that it is a common mutation in catel manzke syndrome . \\
    
    \cmidrule(lr){1-2}
    \small \textbf{CoLo} & \small he required tracheostomy for severe airway hypotonia with pharyngomalacia and laryngomalacia , and gastrostomy tube placement for failure to thrive likely secondary to his pharyngeal dysphagia and airway obstruction . we describe an additional patient with molecularly confirmed catel manzke syndrome who has pierre robin sequence ( without cleft palate ) and manzke dysostosis , and we compare his phenotype with the phenotype of the seven previously described patients with pathogenic variants in tgds . catel manzke syndrome ( mim : 616145 ) is a rare genetic disorder characterized by pierre robin sequence with hyperphalangy and clinodactyly of the index finger . recently implicated homozygous or compound heterozygous pathogenic variants in tgds ( dtdp - d - glucose 4,6-dehydratase ) as the causative factor in a series of seven unrelated patients with features of catel manzke syndrome . of the orthopedic abnormalities are the result of an accessory bone between the second metacarpal and proximal phalanx that causes radial deviation of the second digit ( manzke dysostosis ) . affected individuals typically present with the classic features of pierre robin sequence including micrognathia , airway obstruction secondary to displacement of the tongue base , and often , cleft palate . \\
    
    \cmidrule(lr){1-2}
    \small \textbf{OrderSum} & \small catel manzke syndrome ( mim : 616145 ) is a rare genetic disorder characterized by pierre robin sequence with hyperphalangy and clinodactyly of the index finger . we describe an additional patient with molecularly confirmed catel manzke syndrome who has pierre robin sequence ( without cleft palate ) and manzke dysostosis , and we compare his phenotype with the phenotype of the seven previously described patients with pathogenic variants in tgds . recently implicated homozygous or compound heterozygous pathogenic variants in tgds ( dtdp - d - glucose 4,6-dehydratase ) as the causative factor in a series of seven unrelated patients with features of catel manzke syndrome . of he required tracheostomy for severe airway hypotonia with pharyngomalacia and laryngomalacia , and gastrostomy tube placement for failure to thrive likely secondary to his pharyngeal dysphagia and airway obstruction . tgds encodes a member of the short - chain dehydrogenase / reductase ( sdr ) family ; however the specific function of the protein in humans is unknown . in the present report affected individuals typically present with the classic features of pierre robin sequence including micrognathia , airway obstruction secondary to displacement of the tongue base , and often , cleft palate . \\

\end{longtable}


\end{document}